\documentclass{article}

\usepackage{arxiv}

\usepackage[utf8]{inputenc} 
\usepackage[T1]{fontenc}    
\usepackage{hyperref}       
\usepackage{url}            
\usepackage{booktabs}       
\usepackage{amsfonts}       
\usepackage{nicefrac}       
\usepackage{microtype}      
\usepackage{lipsum}
\usepackage{graphicx}
\usepackage{subfig}
\graphicspath{ {./images/} }

\usepackage[table]{xcolor}

\usepackage{array,multirow,graphicx} 
\usepackage{colortbl}
\usepackage[table]{xcolor} 

\usepackage{setspace}
\usepackage{algorithm}
\usepackage{algorithmic}
\usepackage{subfig}

\title{LCS-DIVE: An Automated Rule-based Machine Learning Visualization Pipeline for Characterizing Complex Associations in Classification}
\author{
Robert Zhang \\
  Institute for Biomedical Informatics\\
  University of Pennsylvania\\
  Philadelphia, PA, 19104 \\
  \texttt{robertzh@wharton.upenn.edu} \\
  \And
Rachael Stolzenberg-Solomon \\
  Division of Cancer Epidemiology and Genetics\\
  National Cancer Institute\\
  Shady Grove, MD, USA \\
  \texttt{rachael.solomon@nih.gov} \\
    \And
Shannon M. Lynch \\
  Cancer Prevention and Control\\
  Fox Chase Cancer Center\\
  Philadelphia, PA, USA \\
  \texttt{shannon.lynch@fccc.edu} \\
    \And
Ryan J. Urbanowicz \\
  Institute for Biomedical Informatics\\
  University of Pennsylvania\\
  Philadelphia, PA, 19104 \\
  \texttt{ryanurb@upenn.edu} \\
}

\begin{document}
\maketitle


\begin{abstract}
Machine learning (ML) research has yielded powerful tools for training accurate prediction models despite complex multivariate associations (e.g. interactions and heterogeneity). In fields such as medicine, improved interpretability of ML modeling is required for knowledge discovery, accountability, and fairness. Rule-based ML approaches such as Learning Classifier Systems (LCSs) strike a balance between predictive performance and interpretability in complex, noisy domains. This work introduces the LCS Discovery and Visualization Environment (LCS-DIVE), an automated LCS model interpretation pipeline for complex biomedical classification. LCS-DIVE conducts modeling using a new scikit-learn implementation of ExSTraCS, an LCS designed to overcome noise and scalability in biomedical data mining yielding human readable `IF:THEN' rules as well as `feature-tracking’ scores for each training sample. LCS-DIVE leverages feature-tracking scores and/or rules to automatically guide characterization of (1) feature importance (2) underlying additive, epistatic, and/or heterogeneous patterns of association, and (3) model-driven heterogeneous instance subgroups via clustering, visualization generation, and cluster interrogation. LCS-DIVE was evaluated over a diverse set of simulated genetic and benchmark datasets encoding a variety of complex multivariate associations, demonstrating its ability to differentiate between them and then applied to characterize associations within a real-world study of pancreatic cancer.
\end{abstract}


\section{Introduction}
Machine learning (ML) plays a ubiquitous role in data science including methodologies for clustering, feature learning, feature importance evaluation, optimization, and predictive modeling (e.g. classification and regression) \cite{jordan2015machine}. In ML predictive modeling there is a recognized trade-off between methods that are regarded as yielding interpretable models (e.g. decision trees \cite{motsinger2010grammatical}), and methods that can achieve higher predictive accuracy (e.g. random forests \cite{qi2012random} and deep feed forward neural networks \cite{najafabadi2015deep}) but are often regarded as `black boxes', i.e. the inner workings are opaque such that even experts cannot fully understand the rationale behind their predictions \cite{carvalho2019machine}. Limited interpretabilty can be a significant drawback in application domains that demand accountability, fairness, and comprehensibility when adopting a prediction machine, and in problems seeking to leverage prediction models to discover knowledge or move beyond identifying associations towards understanding causality \cite{huynh2012statistical,carvalho2019machine,elshawi2020interpretability}.

ML model interpretability can be subjectively as well as objectively evaluated at the global and local levels. Much of the recent ML interpretability research has focused on local interpretability, i.e. explaining individual predictions made by the model, with emphasis on which features (and respective feature values) drove a specific prediction \cite{elshawi2020interpretability}. Local approaches such LIME \cite{ribeiro2016should} or MAPLE \cite{plumb2018model} use secondary analysis of the predictions made by any given ML algorithm, to determine what features were most influential in the given prediction. However, these approaches only estimate this influence, and give minimal insight to the global characterization of a given model, including patterns of association between features and the target outcome prediction. In contrast, global interpretation is viewed as the comprehensibility of the model as a whole outside of understanding a single instance prediction \cite{urbanowicz2012analysis}. This includes the interpretation of (1) globally and locally relevant features (2) their respective value ranges and (3) the nature of the underlying feature associations that drive predictions, i.e. univariate, additive, epistatic, or heterogeneous. In this paper we primarily concern ourselves with global interpretability in noisy biomedical data mining applications. 

Biomedical data can be sourced from various `omics' (e.g. genomics), electronic health records, images, and features extracted from natural text \cite{toga2015big}. Analysis of biomedical data is often complicated by noisy signal, large data dimensionality, multiple classes or quantitative outcomes, missing values, class imbalance, and data heterogeneity (i.e., a mix of feature types - categorical vs quantitative, sources, and value ranges) \cite{urbanowicz2018benchmarking,mirza2019machine}. Additionally, many patterns of association may exist in biomedical data, such as univariate, additive, epistatic, or heterogeneous associations. Univariate associations may be independent or additive with respect to target outcome. Feature interactions, otherwise known as \emph{epistasis} in the context of genomics \cite{moore2016epistasis} yield more than the sum of predictive effects when considering a combination of features simultaneously (e.g. 2-way, 3-way, and beyond). Detection of epistasis among features that have underlying univariate effects (i.e. impure epistasis \cite{urbanowicz2012gametes}), is expected to be easier, while detecting `pure epistatic interactions', i.e. where those features have no univariate effects, is more challenging \cite{urbanowicz2018benchmarking}. A less well-studied complex association, particularly relevant to this work, is that of \emph{heterogeneous associations}, otherwise known as genetic heterogeneity in genomics \cite{ritchie2003power,thornton2006dissecting}. Heterogeneous associations occur when individual (or sets of) features are independently predictive of the same outcome variable. Here, there can be subgroups of instances that are best predicted by a distinct subset of features. Such heterogeneity can be indicative of distinct disease subtypes \cite{dahl2020genetic}. 

Learning classifier systems (LCSs) \cite{urbanowicz2009learning,urbanowicz2017introduction} are rule-based ML algorithms capable of detecting complex multivariate associations (i.e. epistasis and genetic heterogeneity) in both clean and noisy data \cite{urbanowicz2010application,urbanowicz2014extended,urbanowicz2015exstracs}. They yield models comprised of individually interpretable IF:THEN rules from which underlying associations can be characterized, i.e. relevant features, patterns of association, and identification of heterogeneous subgroups \cite{urbanowicz2012analysis,urbanowicz2012instance,urbanowicz2013role,urbanowicz2018attribute}. ExSTraCS was developed as an LCS to improve supervised learning performance and interpretation in noisy, larger-scale biomedical data \cite{urbanowicz2015exstracs}. Among other heuristics, ExSTraCS utilizes a \emph{feature-tracking} strategy that learns which features contribute the most to accurate predictions of each instance during training \cite{urbanowicz2012instance,urbanowicz2018attribute}. Post-training clustering of feature-tracking scores was able to reveal underlying heterogeneous instance subgroups without making prior assumptions of data homogeneity via data stratification \cite{urbanowicz2012instance,urbanowicz2013role}. Previous work has demonstrated that global patterns of feature associations could similarly be characterized via a mixture of statistical analyses, clustering, and various visualizations of the rules themselves \cite{urbanowicz2012analysis,liu2021visualizations}. However, to date, this process has been largely ad hoc, and the capabilities and limitations of these ExSTraCS-based interpretation strategies for complex, noisy problems has not been comprehensively addressed towards making further advancements. We hypothesize that the feature-tracking scores generated by ExSTraCS along with global rule-set analysis can be leveraged to characterize and distinguish unique patterns of association (i.e. univarate, additive, epistatic, and heterogeneous) and automatically identify candidate subgroups of instances reflecting underlying heterogeneity.

In this paper we first introduce scikit-ExSTraCS, a new accessible implementation of the ExSTraCS algorithm. Next, we introduce LCS-DIVE (LCS Discovery and Visualization Environment), an automated LCS model interpretation pipeline for noisy, complex classification problems. LCS-DIVE conducts modeling with scikit-ExSTraCS, which yields rules and feature-tracking scores that are used to discover and visualize (1) feature importance (2) underlying patterns of association, and (3) model-driven candidate heterogeneous instance subgroups. We demonstrate the capabilities and limitations of LCS-DIVE for model interpretation via application to a range of simulated data scenarios depicting different patterns of association, numbers of predictive features, and degrees of noise. We emphasize the ability to utilize feature-tracking scores to characterize heterogeneous associations and correctly recover respective instance subgroups. We apply this pipeline and simulation study findings to facilitate the interpretation of a real-world investigation of pancreatic cancer.  


\section{Methods}
In this section we briefly review LCS algorithms, introduce scikit-ExSTraCS, and further explain feature-tracking (FT). Next, we detail the LCS-DIVE pipeline using benchmark multiplexer problems to illustrate this process, i.e. automated clustering and visualization of both FT scores and rule-sets. Lastly, we summarize the simulated and real-world pancreatic cancer datasets that comprise our pipeline evaluations.

\subsection{The scikit-ExSTraCS package}
Learning classifier systems (LCSs) are a family of rule-based machine learning (RBML) algorithms where model discovery is driven by an evolutionary algorithm (EA) component. LCSs are most distinguished from other EAs and ML predictive modeling approaches based on how models are represented in a `piece-wise’ manner. Specifically, a number of conditional IF:THEN rules (e.g. IF age $>$ 15 AND height $>$ 6 THEN male) are discovered and applied as a conditional ensemble to make predictions. Each rule can either specify or ignore any of the features in the dataset. In predicting a given instance, only rules that have a condition satisfied (i.e. matched) by the respective instance values are applied from the learned rule set. This characteristic makes LCS algorithms uniquely well suited to capturing heterogeneous associations in addition to univariate and epistatic effects \cite{urbanowicz2015exstracs}. LCSs are best suited to classification tasks, however they have also been adapted to regression problems as well \cite{iqbal2012xcsr,urbanowicz2015continuous}. The earliest and best known LCS algorithms were designed for reinforcement learning, e.g. the Michigan-style LCS known as XCS \cite{wilson1995classifier}. Some LCS variations have since been specialized to supervised learning, e.g. UCS \cite{bernado2003accuracy} and eLCS \cite{urbanowicz2017introduction}, directly descended from XCS. Further introduction to LCS algorithms are available \cite{urbanowicz2009learning,urbanowicz2017introduction}. ExSTraCS \cite{urbanowicz2014extended} expanded upon the UCS framework, integrating FT (described in detail below) \cite{urbanowicz2012instance}, expert knowledge guided evolutionary rule discovery \cite{urbanowicz2012using}, and rule compaction, which seeks to remove inexperienced or redundant rules from the rule population post-training \cite{tan2013rapid}. Later, `ExSTraCS 2.0' added a rule-specificity limit and efficient rule representation to improve scalability in datasets with a larger number of features and more complex underlying multivariate associations \cite{urbanowicz2015exstracs}.

Previous ExSTraCS implementations did not follow the standard interfaces of modern ML libraries. This made them less convenient to use within a ML analysis pipeline for application and comparison to other established ML methods. Using the recently developed scikit-eLCS \cite{zhang2020scikit} as a blueprint, here we developed scikit-ExSTraCS, a scikit-learn compatible near-replicate of the ExSTraCS 2.0 algorithm \cite{urbanowicz2015exstracs}. Scikit-ExSTraCS has the same interface as standard algorithms in the popular scikit-learn Python library. It includes functionality for setting hyperparameters, applying the trained model as a prediction model, and retrieving key output files (e.g. including the rule population and FT scores). Scikit-ExSTraCS has four differences from ExSTraCS 2.0 \cite{urbanowicz2015exstracs} : (1) correct set subsumption was disabled based on previous findings \cite{lanzi2007empirical}, (2) addition of an improved FT approach \cite{urbanowicz2018attribute}, (3) expert knowledge scores were generated using the superior MultiSURF feature importance estimation algorithm \cite{urbanowicz2018benchmarking} but without the TuRF wrapper \cite{moore2007tuning}, and (4) while included in the scikit-ExSTraCS software, for simplicity, rule compaction was not applied in most experimental evaluations since it does not impact FT scores. The scikit-ExSTraCS code and documentation are available at: \cite{scikit_exstracs}. Scikit-ExSTraCS is a stand-alone package that can be used outside of LCS-DIVE to generate predictive ML models without seeking a deeper interpretation.

\subsubsection{Feature Tracking in scikit-ExSTraCS}
Feature tracking (FT) \cite{urbanowicz2012instance} is an instance-level memory mechanism for Michigan-style supervised LCSs such as UCS \cite{bernado2003accuracy}, eLCS \cite{urbanowicz2017introduction,zhang2020scikit}, and ExSTraCS\cite{urbanowicz2015exstracs}. Previously, this mechanism was referred to as `attribute tracking', however `feature tracking' is better in-line with ML terminology, e.g. feature selection \cite{urbanowicz2012instance,urbanowicz2014extended,urbanowicz2015exstracs}. FT serves two purposes: (1) during training, accumulated FT scores are probabilistically applied as a source of experiential knowledge to better guide evolutionary rule discovery \cite{urbanowicz2012instance}, and (2) these scores can be clustered and analyzed 'post-training' to characterize patterns of association and heterogeneous instance subgroups. Analyzing FT scores is the primary focus of the LCS-DIVE pipeline. 

Scikit-ExSTraCS adopts an improved Widrow-Hoff (time-weighted recency) FT update strategy detailed in \cite{urbanowicz2018attribute}. The Widrow-Hoff update places more weight on rules surviving in the population later in model training (i.e. when a more successful set of rules has ideally evolved). This new FT scoring did little to impact LCS predictive performance, but it was found to significantly improve model interpretability \cite{urbanowicz2018attribute}. Each learning iteration, scikit-ExSTraCS focuses on a single training instance. A match set [M] is formed by all existing rules that matches this instance, and subsequently a correct set [C] is formed including all rules in [M] with the correct class. For any feature that has a 'specified' value in a rule of [C], FT scores are updated with the Widrow-Hoff equation ($\beta = 0.1$) for the current training instance.

\begin{equation} \label{eq:WH}
Ave_{New} = Ave_{Current} + \beta(Value_{Current} - Ave_{Current})
\end{equation}

The resulting FT scores have the same dimensionality as the original dataset. Model training will generate a FT signature for each training instance that reflects the features most important to it's correct classification (i.e. those with a larger FT score). Since rules are highly adept at capturing both univariate and epistatic effects, FT scores are expected to reflect those patterns as well. Ultimately, the goal is to cluster instances with similar FT signatures as a homogeneous subgroup within a heterogeneous problem.

\subsection{The LCS-DIVE Pipeline}
\begin{figure}[t]
  \includegraphics[width=1.0\linewidth, keepaspectratio]{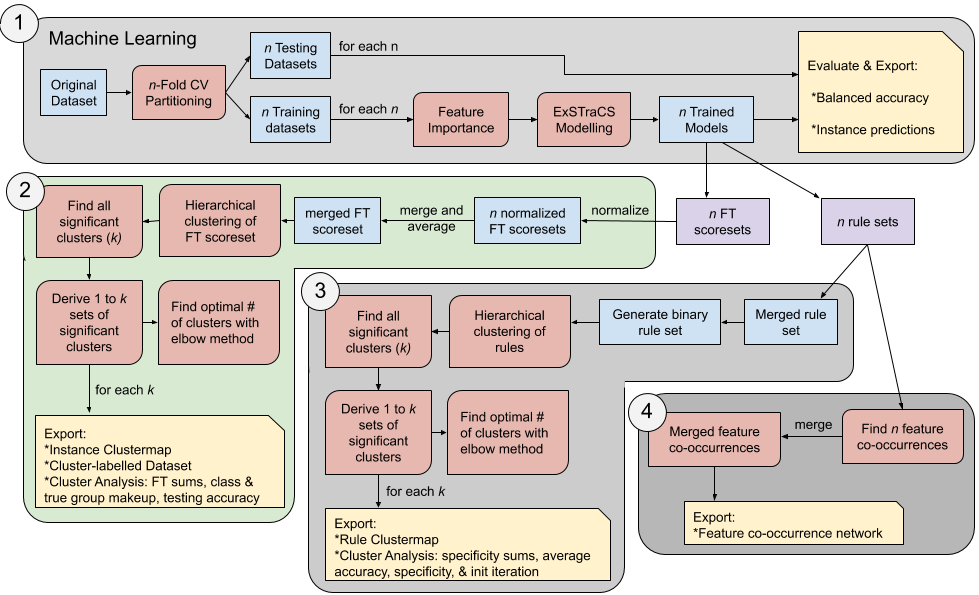}
  \caption{Schematic of the LCS-DIVE pipeline. (1) scikit-ExSTraCS ML modeling (2) FT significant clustering, heatmap generation and significant instance subgroup identification, (3) rule population heatmap generation (4) rule population co-occurence network generation}
  \label{fig:Pipeline}
\end{figure}

The LCS-DIVE pipeline is summarized by Figure \ref{fig:Pipeline}. The pipeline is comprised of 4 major phases: (1) ML modeling with scikit-ExSTraCS (2) FT score interpretation via clustering, heatmap visualization, and subgroup identification, (3) rule-based model interpretation via clustering, heatmap visualization, and subgroup identification, and (4) rule-based model interpretation via co-occurence network generation. LCS-DIVE was coded in Python 3 and can be run locally from the command line but also on a parallel CPU computing environment. The open source code is available here: \cite{lcs_dive}. 

\subsubsection{MUX Example Benchmark Problem} \label{mux}
To illustrate the LCS-DIVE phases below, we present an example application to the 11-bit Multiplexer (MUX) binary classification problem which includes 11 relevant bits (i.e. features), that comprise clean (i.e. no noise) patterns of both pure epistasis and heterogeneity. MUX problems are longstanding ML benchmarks often examined in LCS algorithm research \cite{wilson1987classifier,urbanowicz2015exstracs}. They are easily scaled up to examine increasingly complex patterns of association in larger feature spaces. In the 11-bit MUX, three 'address' bits (i.e. A0, A1, A2) point to a specific 'register' bit (e.g. R5), the value of which specifies the true class. To determine the class of a given instance, the values of the three address bits, and the unique register bit to which they point are required as a pure 4-way feature interaction. Since the values of the address bits determine what register bit is relevant, a distinct subset of features is required to correctly predict distinct subgroups of instances in the dataset (i.e. a heterogeneous association).

\subsubsection{Phase One: Modeling With scikit-ExSTraCS}
Phase one of LCS-DIVE applies ML modeling with scikit-ExSTraCS. \emph{n}-fold cross validation (CV) is applied yielding training and testing sets. An \emph{n} of 10 was used across this study. For each training dataset, feature importance scores are first estimated with MultiSURF \cite{urbanowicz2018benchmarking}, a filter-based feature importance estimation algorithm demonstrated to effectively detect univariate, epistatic and heterogeneous associations. These scores are passed to scikit-ExSTraCS as a training parameter and used as expert knowledge to accelerate optimal rule discovery \cite{urbanowicz2012using}. Post training, respective testing datasets are used to evaluate model performance via balanced accuracy. See Sup.1.1 for details on data formatting and Sup.1.2 for details on scikit-ExSTraCS hyperparameter settings Supplementary Materials available at: \cite{lcs_dive}.

\subsubsection{Phase Two: Feature Tracking Interpretation}
Phase two of LCS-DIVE focuses on FT score interpretation across each of the \emph{n}-fold CV training datasets. This phase is the primary focus of LCS-DIVE and includes the following steps: FT normalization, merging across CV partitions, hierarchical clustering, significant cluster identification, clustermap generation, and optimal cluster number determination. This phase yields clustermaps of significant instance subgroups (i.e. clusters) within the original dataset and a recommendation for the `optimal' number of clusters. Clusters represent instance subsets with homogeneous FT feature importance signatures. The approach laid out here is an extended and automated departure from previous applications of FT \cite{urbanowicz2012instance,urbanowicz2013role,urbanowicz2015exstracs}.

\begin{figure}[tph]
    \centering
    \subfloat[\centering Unclustered Heatmap]{{\includegraphics[width=0.45\linewidth, keepaspectratio]{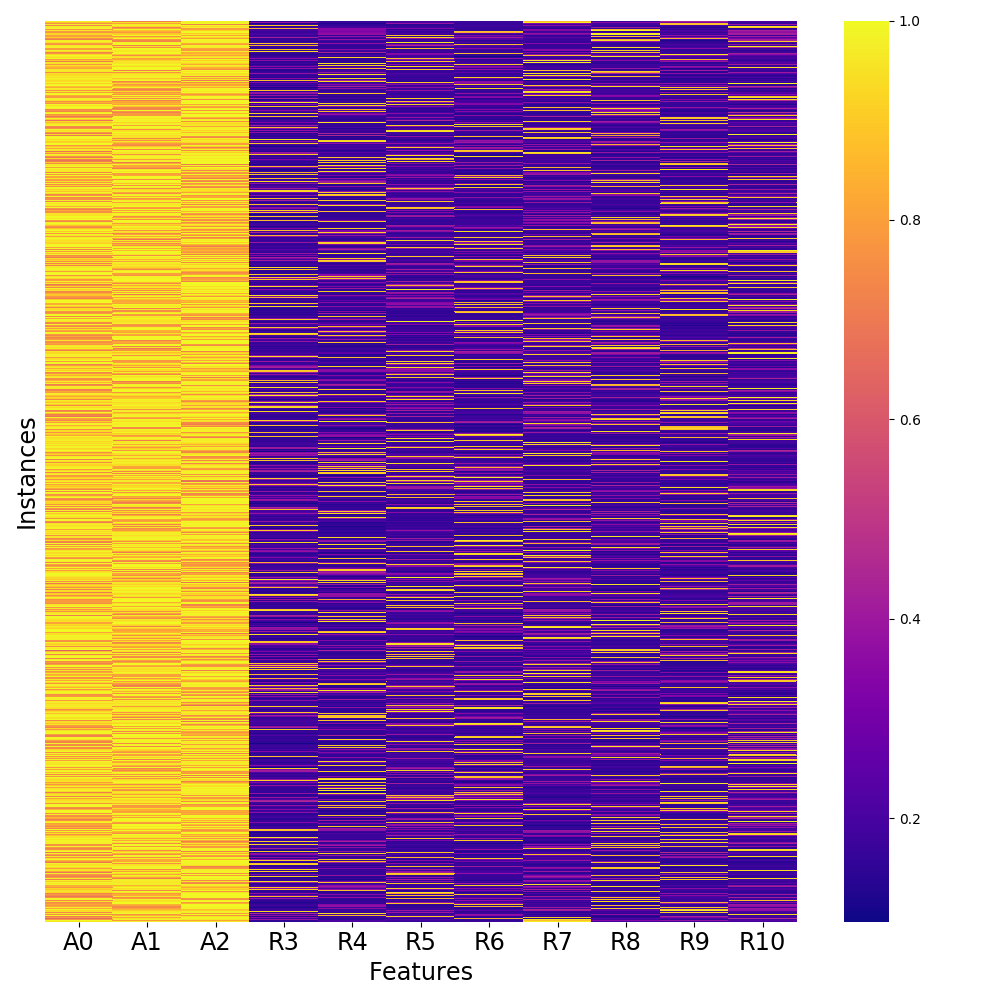}}}%
    \qquad
    \subfloat[\centering Clustered Heatmap]{{\includegraphics[width=0.45\linewidth, keepaspectratio]{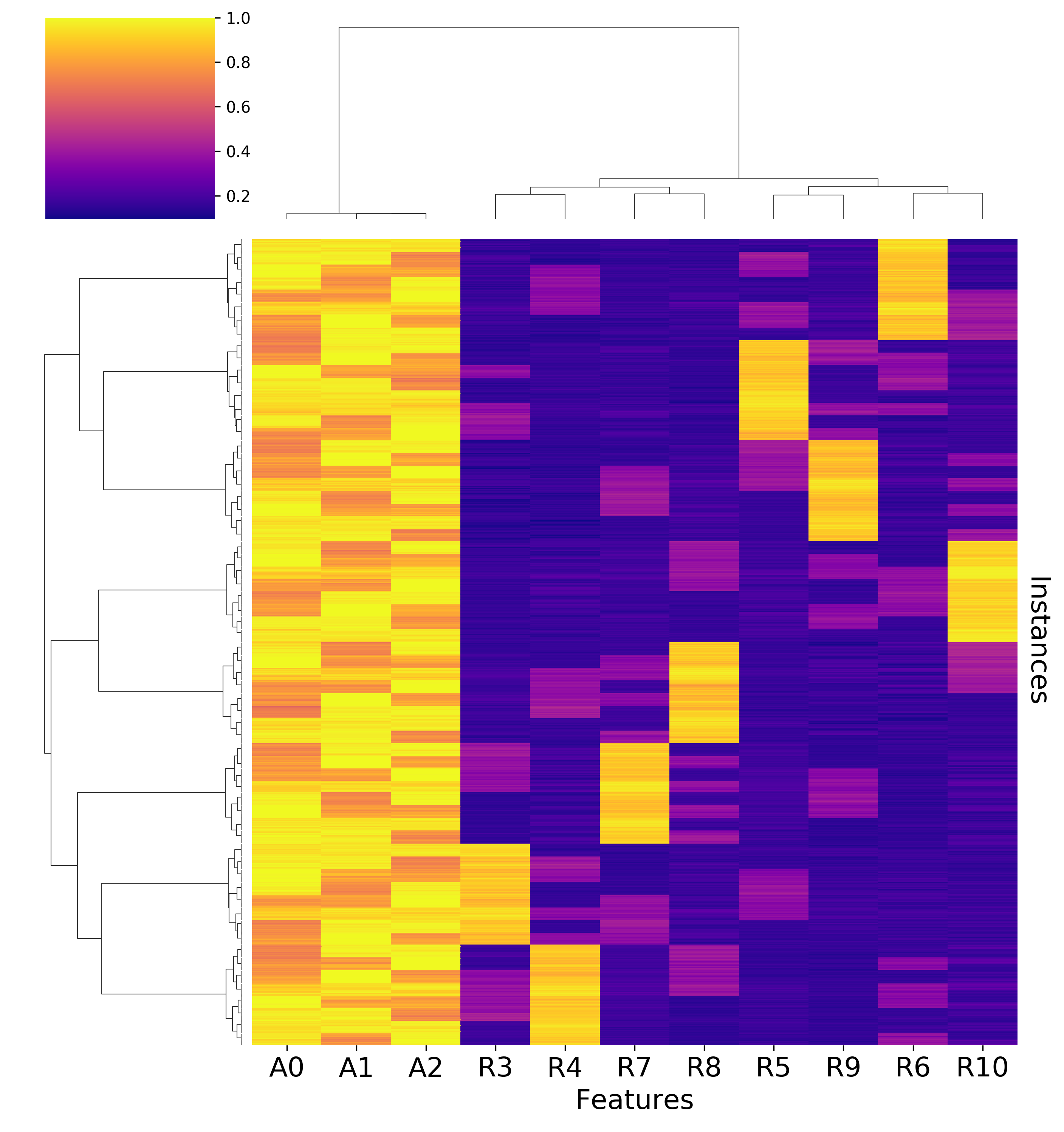}}}%
    \caption{LCS-DIVE heatmap (A) and clustermap (B) of FT scores from the 11-bit MUX problem}
    \label{fig:ATClustermap_unlabelled}
\end{figure}

The first step normalizes each of the \emph{n} FT score sets independently, i.e. dividing the scores of each instance by that instance’s maximum FT score, transforming all scores between 0 to 1. This normalization accounts for the fact that `optimal' rules to solve the target problem may be discovered at different time points during model training, and thus have different opportunities to impact the magnitude of FT scoring within different instances \cite{urbanowicz2012instance}. The second step merges the \emph{n} CV normalized FT scores by averaging the (\emph{n}-1) FT score rows where a given instance appeared in a CV training dataset. This yields a single set of FT scores with the dimensionality of the original dataset representing average FT scores for each instance across \emph{n}-fold CV modeling. The third step hierarchically clusters the normalized and merged FT scores. Hierarchical clustering was chosen over methods like k-means clustering \cite{kaushik2014hvk}, to avoid assumptions of the target number of clusters. Applying the `Seaborn' package, Pearson correlation was used to calculate distance between instances and features and the Ward method was used to find the distance between hierarchical clusters \cite{waskom2020seaborn}. See Sup.1.3.1 for further clustering details. Using the 11-bit MUX example, Figure \ref{fig:ATClustermap_unlabelled} illustrates normalized and merged FT scores before and after clustering. Prior to clustering, it's clear that that the three address bits (i.e. A0, A1, A2) are uniformly important to predictions across all instances. Post-clustering the feature subgroups of address bits and their corresponding register bit become apparent. Further, 8 distinct instance subgroups appear, each with 4 features with high FT scores. While the ideal number of clusters in this example is clear, this will likely not be the case in noisy problems. This will be addressed by the subsequent steps. 

The fourth step seeks to identify the smallest clusters of instances that can be considered significant, i.e. how far down the dendrogram we can travel before the distance between individual clusters are no longer statistically significant ($< 0.05$). This will yield some maximum number of significant clusters (\emph{k}) that can quite likely be larger than the optimal number of meaningful clusters. For example, LCS-DIVE found a maximum of 52 significant clusters for the 11-bit MUX, when the optimal number of clusters is 8. To determine statistically significant clusters we implemented a previously proposed Monte Carlo based method \cite{kimes2017statistical}. See Sup.1.3.2 for details on this approach and cluster outputs generated by LCS-DIVE.

\begin{figure}[tph]
    \centering
    \subfloat[\centering Clustermap]{{\includegraphics[width=0.45\linewidth, keepaspectratio]{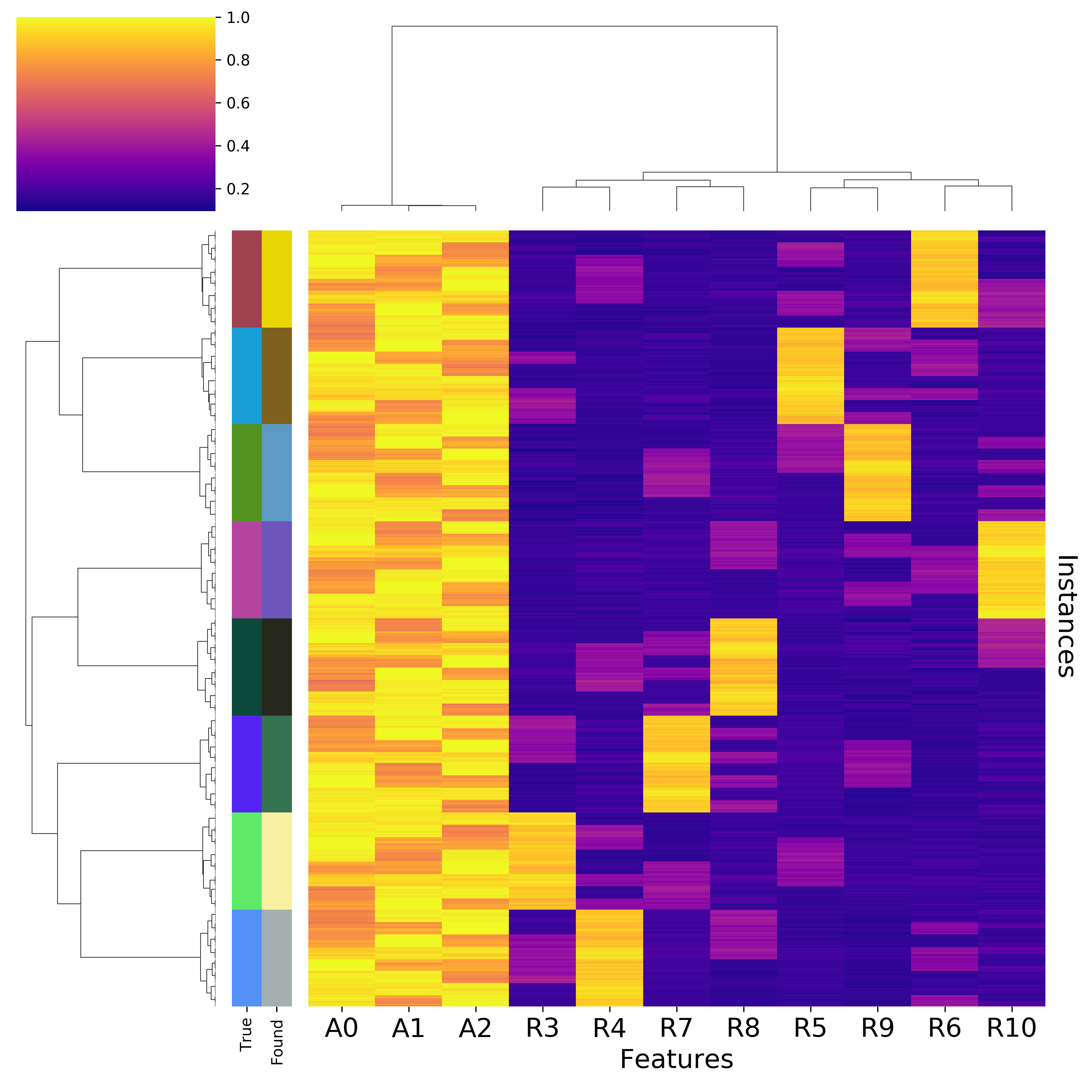}}}%
    \qquad
    \subfloat[\centering Elbow Plot]{{\includegraphics[width=0.45\linewidth, keepaspectratio]{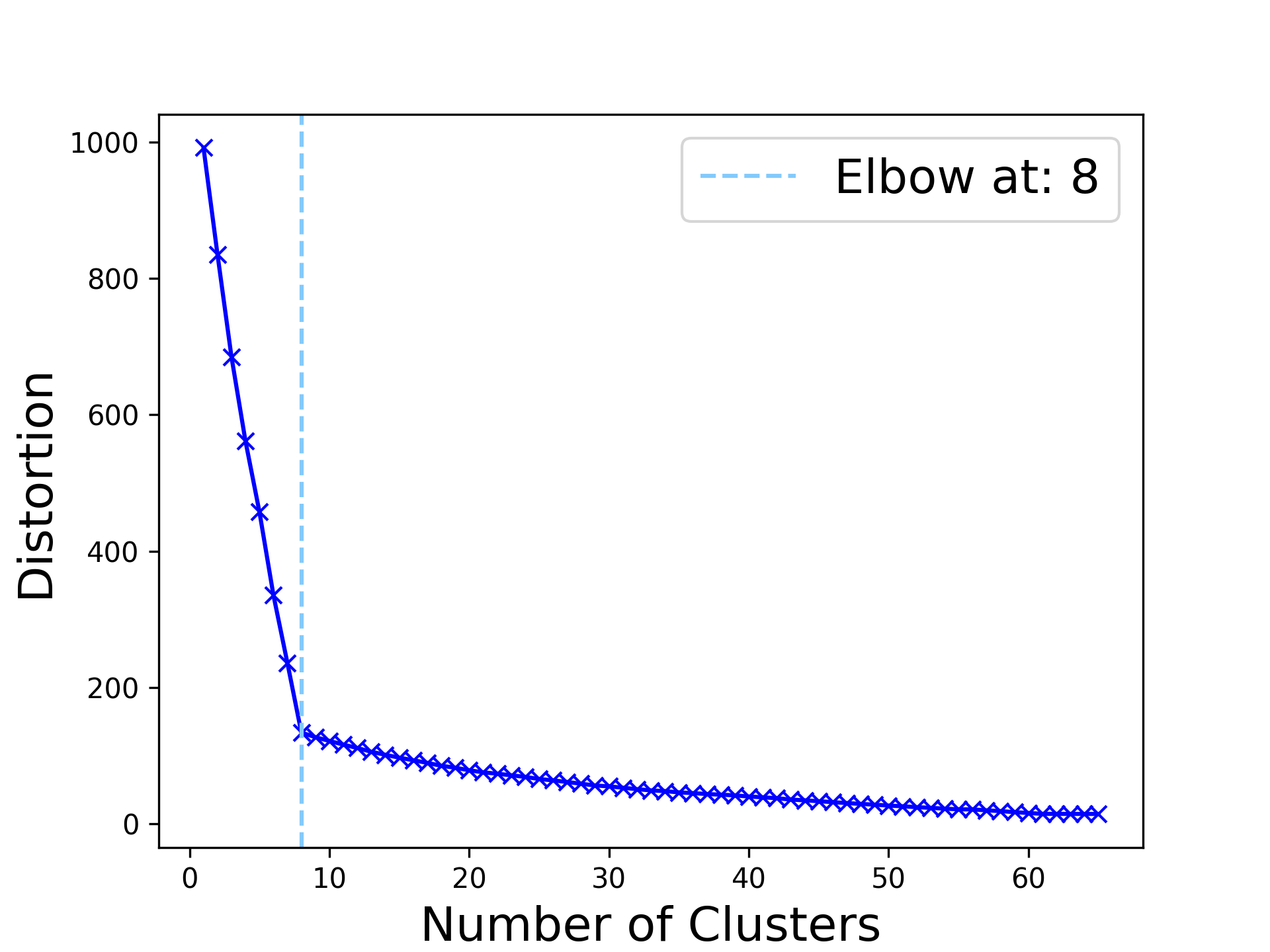}}}%
    \caption{FT score visualization of 11-bit MUX problem (A) Significant clustermap (\emph{k} $= 8$) including 'found' and 'true clusters and (B) the corresponding elbow plot}
    \label{fig:ATSample}
\end{figure}

The fifth step produces clustermaps for all possible significant cluster counts, i.e from 1 homogeneous cluster (encompassing all instances) to \emph{k} maximum significant clusters. These can be viewed as distinct clustering hypotheses in search of the optimum. Generating candidate clustermaps from \emph{k} to 1 is achieved by moving up the dendrogram: merging the closest pair of clusters (i.e. the shortest Ward distance) that also share the same direct parent cluster. This is repeated step-wise until \emph{k} clustermaps have been generated. Each clustermap can include up to two additional columns: (1) cluster membership labels for each instance and (2) the true underlying cluster membership for each instance, if known and provided by the user, for downstream comparison. Figure \ref{fig:ATSample}A gives the significant clustermap (\emph{k} $= 8$) for the 11-bit MUX including columns for the 'found' clusters, and 'true' underlying clusters. We can see in this example that the found clusters perfectly align with the true ones. For each clustermap, LCS-DIVE also exports (1) the original dataset, adding a 'clusterID' column such that followup analyses may be conducted linking instances in the data to identified clusters, and (2) a spreadsheet summarizing cluster analysis characteristics such as `within-cluster' testing accuracy. 

The last step seeks to facilitate determination of the ‘optimal’ number of homogeneous clusters. LCS-DIVE employs the elbow method with the distortion metric (i.e. the sum of square distances from each point in a cluster to it's centroid) \cite{rohmandistortion2018}. The elbow approach seeks to pick some minimal number of clusters which also minimizes distortion. One elbow plot is generated describing the set of \emph{k} possible significant clustermaps. Importantly, LCS-DIVE makes all \emph{k} clustermaps available so that investigators have the option to use their own strategy or judgement for selecting the optimal number of clusters or to compare different candidate heterogeneous subgroup clusters downstream. Figure \ref{fig:ATSample}B gives the elbow plot generated for the 11-bit MUX. LCS-DIVE attempts to automatically identify the elbow (i.e optimal cluster count) by rotating the elbow plot counterclockwise about the origin such that the left and right-most points on the curve are level on the y-axis. The selected elbow is the point with the lowest y coordinate. For the 11-bit MUX this approach correctly identifies the optimal 8 instance clusters as seen in Figure \ref{fig:ATSample}B. However this strategy is not guaranteed to be optimal in all scenarios.

\subsubsection{Phase Three: LCS Rule Set Interpretation}
The third phase of LCS-DIVE differently focuses on clustering and visualizing the set of LCS rules constituting the trained model directly. The goal is to interrogate global feature patterns found across the rules of the model to identify important features and infer underlying patterns of association. This phase formalizes and automates the global rule visualization pipeline first proposed in \cite{urbanowicz2012analysis}. First, we take the \emph{n} trained rule populations and merge them into a single set. This merged rule set is not meant to be applied as a predictive model (due to risk of overfitting) without validation of its performance outside of the original dataset. The second step prepares the rule set for visualization by encoding them based on what features were specified vs. ignored (i.e. generalized) within each rule. Rule encoding is detailed Sup.1.4.

\begin{figure}[tph]
    \centering
    \subfloat[\centering Rule Clustermap w/o Compaction]{{\includegraphics[width=0.45\linewidth, keepaspectratio]{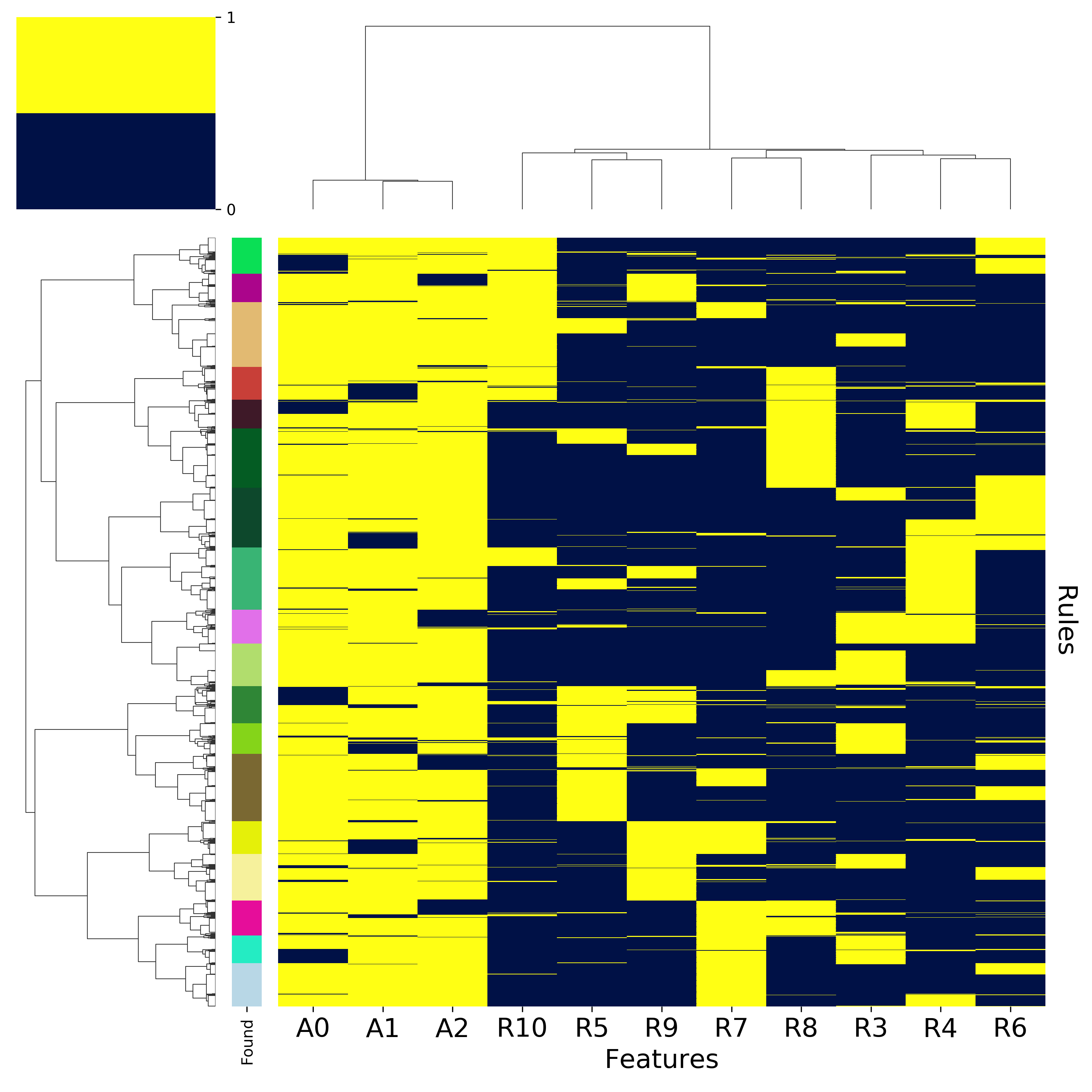}}}%
    \qquad
    \subfloat[\centering Elbow Plot w/o Compaction]{{\includegraphics[width=0.45\linewidth, keepaspectratio]{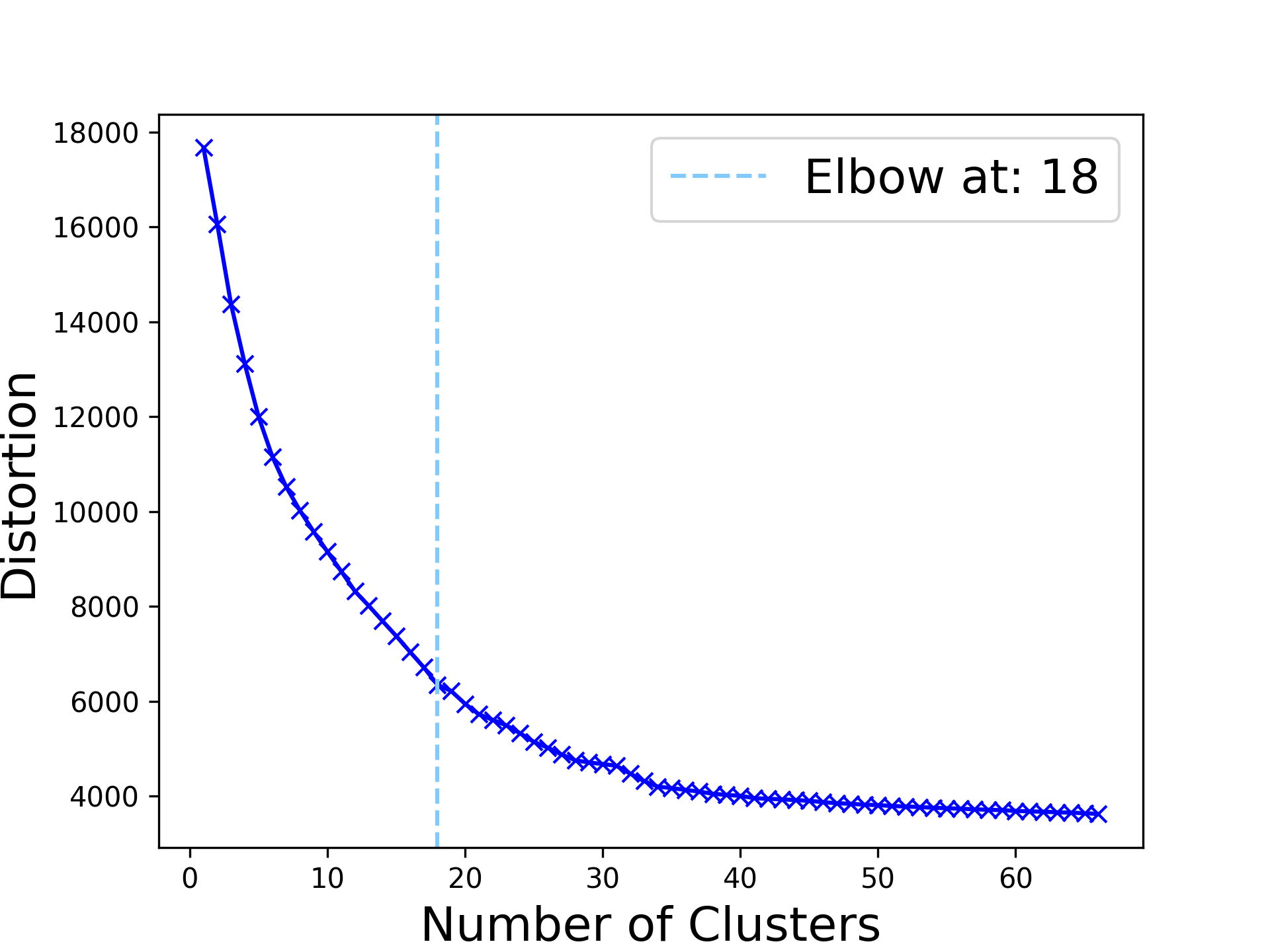}}}%
    \qquad
    \subfloat[\centering Rule Clustermap with Compaction]{{\includegraphics[width=0.45\linewidth, keepaspectratio]{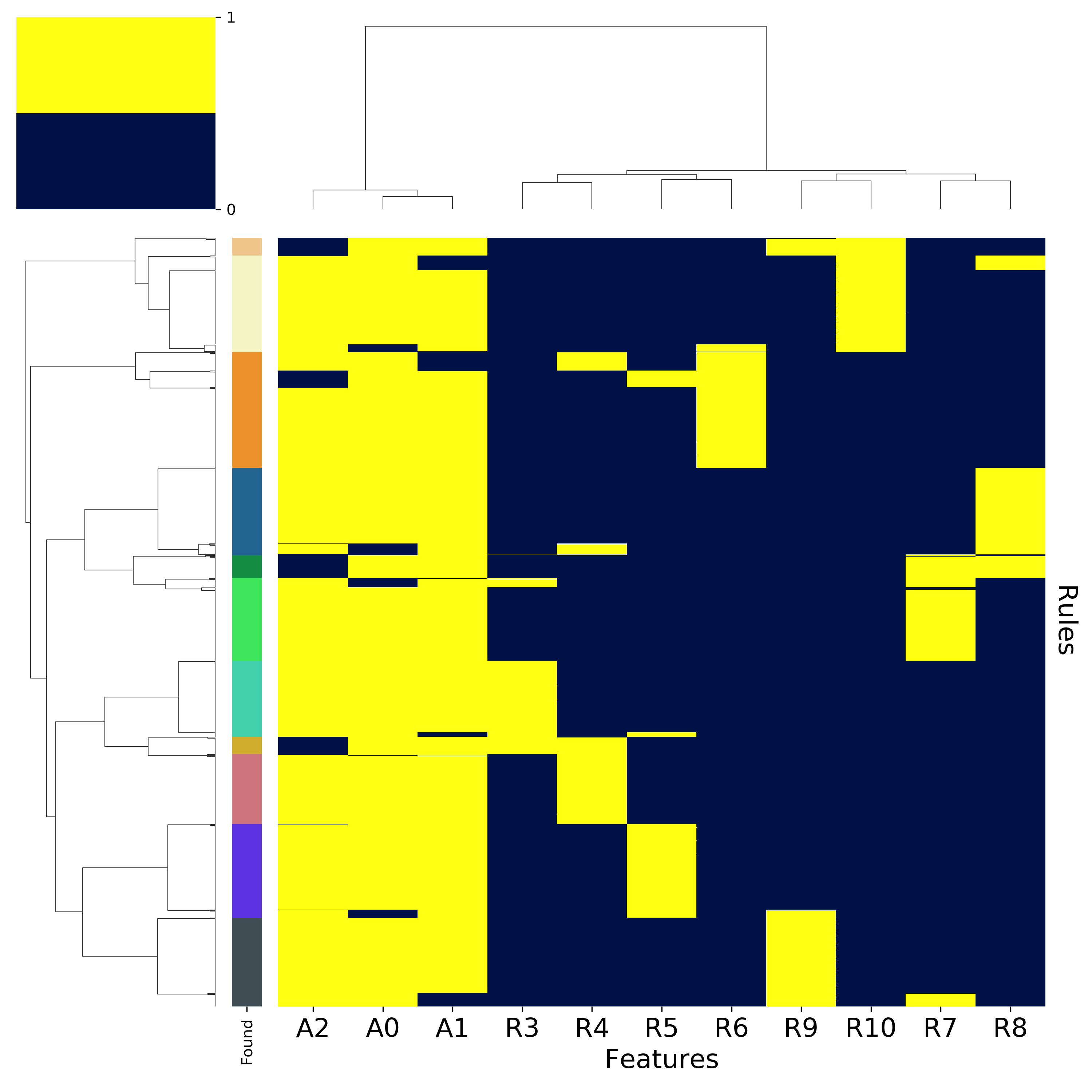}}}%
    \qquad
    \subfloat[\centering Elbow Plot with Compaction]{{\includegraphics[width=0.45\linewidth, keepaspectratio]{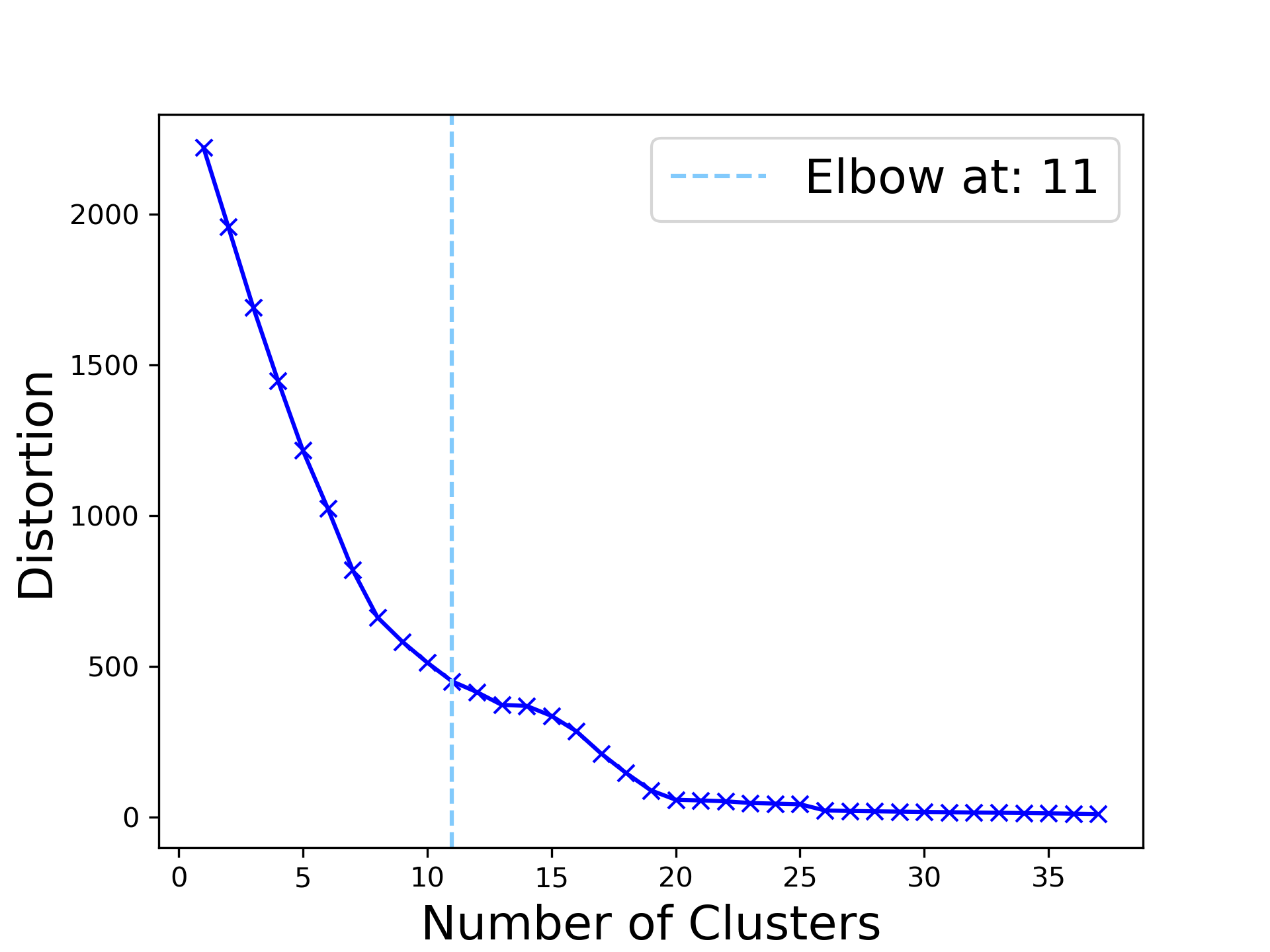}}}%
    \caption{Rule set visualization of 11-bit MUX problem: (A) significant clustermap w/o compaction (\emph{k} $= 18$) including 'found' clusters (B) elbow plot for analysis w/o compaction (C) significant clustermap with compaction (D) elbow plot for analysis with compaction}
    \label{fig:RuleSample}
\end{figure}

The subsequent steps mirror those from phase two including hierarchical clustering, identification of \emph{k} significant clusters, generation of heatmaps, and suggested cluster number determination. Figure \ref{fig:RuleSample} gives the results of clustering the rule set trained on the 11-bit MUX, with and without post-training rule compaction (using CRA2 \cite{dixon2002cra2}). There are no `true' clusters here since we are clustering rules rather than instances. Note that with rule compaction a much cleaner set of rule patterns emerge similar to those observed in Figure \ref{fig:ATSample} using FT scores. Further, a cluster number closer to the ideal 8 feature combination clusters is recommended by the elbow plot approach. Rule compaction has no impact on FT scores that are generated during model training, thus it is not applied in subsequent analyses of this paper. However, it is an important element for improving LCS rule population interpretation as illustrated here \cite{tan2013rapid,liu2019absumption}.

\subsubsection{Phase Four: LCS Rule Co-occurence Network Interpretation}
\begin{figure}[tph]
    \centering
    \subfloat[\centering Network w/o compaction]{{\includegraphics[width=0.45\linewidth, keepaspectratio]{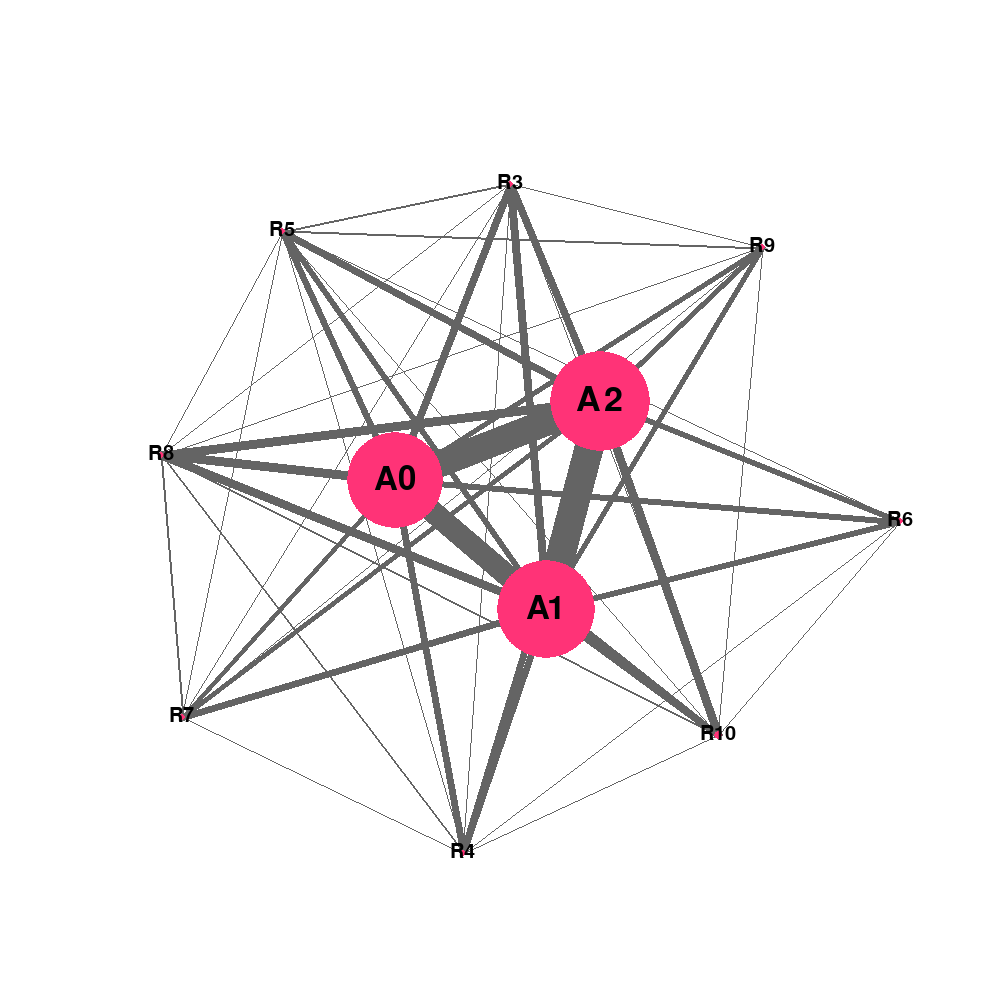}}}%
    \qquad
    \subfloat[\centering Network with compaction]{{\includegraphics[width=0.45\linewidth, keepaspectratio]{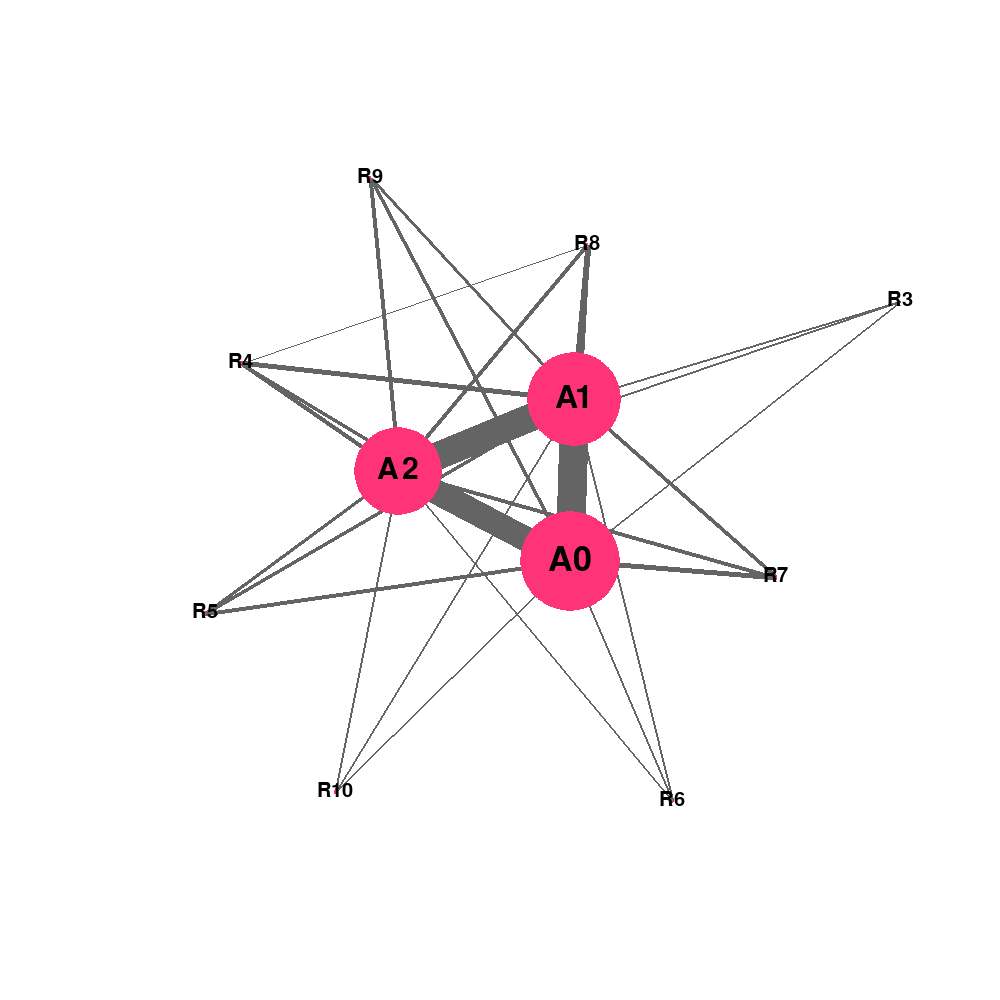}}}%
    \caption{Feature Co-occurrence Network for the 11-bit MUX (A) with and (B) without CRA rule compaction. Nodes are features where the diameter is the specification sum across all rules. Edges correspond to co-specification of a pair of features across all rules}
    \label{fig:networkSample}
\end{figure}

 The final phase of LCS-DIVE also operates on the merged rule set, generating a network visualization. The goal here is to further characterize underlying associations by visualizing univariate, epistatic, and heterogeneous patterns captured in the rule set \cite{urbanowicz2012analysis}. First, we sum pairwise specified feature co-occurences across the merged binary encoded rule set. Co-specification in rules indicates potential feature interactions \cite{urbanowicz2012analysis}. Then a network visualization is generated where nodes are dataset features and node diameter is proportional that features specification across all rules. Edge thickness between nodes represents pairwise co-specification counts of those features across all rules. Figure \ref{fig:networkSample} gives the network generated for the 11-bit MUX with and without CRA2 rule compaction. A clear interaction between the three address bits is seen, and respective (less frequent) heterogeneous interactions between address and respective register bits is also observed. Notice that co-occurences between register bits are negligible (made clearer following rule compaction), offering support in properly characterizing the underlying epistatic and heterogeneous associations underlying the 11-bit MUX.
 
\subsection{Data and Experimental Evaluations} \label{data_sec}
To evaluate LCS-DIVE we first applied it to a broader spectrum of clean MUX benchmark problems to demonstrate the interpretable power of LCS-DIVE in unambiguous, yet complex problems. The 6, 11, 20, 37, and 70-bit MUX problems were simulated as benchmark datasets as described in \cite{urbanowicz2015exstracs} and detailed in Sup.2.1. Table \ref{tab:muxparams} gives the characteristics of each dataset as well as the hyperparameters used by MultiSURF and scikit-ExSTraCS.  

\begin{table}[tph]
\centering
\captionof{table}{\emph{x}-MUX benchmark datasets and analysis hyperparameters where `a' is the number of address bits, 'N' is the maximum rule set size, \emph{nu} is the exponential emphasis placed on rule accuracy \cite{urbanowicz2015exstracs}, and the MultiSURF 'sample' hyperparemeter refers to the random sample of training instances used to estimate feature importance scores} \label{tab:muxparams} 
\begin{tabular}{|c|c|c|c|c|c|c|c|c|}
\hline
\multicolumn{5}{|c|}{\textbf{MUX}} & \multicolumn{4}{|c|}{\textbf{Hyperparameters}} \\
\hline
\multicolumn{5}{|c|}{\textbf{Dataset Characteristics}} & \multicolumn{3}{|c|}{\textbf{scikit-ExSTraCS}} & \multicolumn{1}{|c|}{\textbf{MultiSURF}} \\
\hline
& \textbf{a} & \textbf{m-way} & \textbf{Heterogeneous} & \textbf{Number of} & \textbf{Learning} &  &  &  \\
\textbf{x-MUX} & & \textbf{Interactions} & \textbf{Combinations} & \textbf{Instances} & \textbf{Iterations} & \textbf{N} & \textbf{nu} & \textbf{Samples} \\
\hline
6-bit    & 2 & 3 & 4 & 500 & 20k & 500 & 10 & All\\
\hline
11-bit   & 3 & 4 & 8 & 1000 & 20k & 2000 & 10 & All\\
\hline
20-bit   & 4 & 5 & 16 & 2000 & 100k & 5000 & 10 & 1000      \\       
\hline
37-bit   & 5 & 6 & 32 & 5000 & 200k & 5000 & 10 & 1000    \\       
\hline
70-bit   & 6 & 7 & 64 & 10k & 500k & 10000 & 10 & 9000     \\   
\hline
\end{tabular}
\end{table}

Additionally we evaluated LCS-DIVE across a variety of simulated single nucleotide polymorphism (SNP) datasets with different clean vs. noisy signals, and distinct underlying patterns of association, i.e. univariate, additive, epistatic, heterogeneous, and mixed. The GAMETES software \cite{urbanowicz2012gametes} for simulating complex single nucleodide polymorphism SNP data was applied to generate a variety of unique complex association scenarios as detailed in Sup.2.2. A total of 21 unique datasets were simulated and analyzed here to illustrate and evaluate how LCS-DIVE can differentiate interpretation of underlying associations. The characteristics of these datasets are summarized in Table \ref{tab:simdata}. For simplicity, all datasets were simulated with 20 features and balanced proportions of binary case/control outcomes. Model heritability indicates the proportion of signal in the data, i.e. a value of 1 is clean (no noise). Clean, pure epistatic datasets (i.e. 7-10) were generated with a 2-way or 3-way XOR model as in \cite{urbanowicz2018benchmarking}. 

\begin{table}[tph]
\centering
\captionof{table}{Characteristics of the 21 GAMETES Simulated Datasets} \label{tab:simdata}
\begin{tabular}{|c|c|c|c|c|c|c|}
\hline
\textbf{Dataset} & & \textbf{Predictive} & \textbf{$\#$ of} & \textbf{Model} & \textbf{Model} &  \\
\textbf{ID} & \textbf{Underlying Association}  & \textbf{Features} & \textbf{Models} & \textbf{Ratio} & \textbf{Heritability} & \textbf{Instances} \\
\hline
1 $\&$ 2  & univariate & 1 & 1 & NA & 1 $\&$ 0.4 & 1600\\ \hline
3 $\&$ 4 & additive & 2 &  2 & equal & 1 $\&$ 0.4 & 1600\\ \hline
5 $\&$ 6 & additive & 4 &  4 & equal & 1 $\&$ 0.4 & 1600\\ \hline
7 $\&$ 8 & 2-way pure epistasis & 2 &  1 & NA & 1 $\&$ 0.4 & 1600\\ \hline
9 $\&$ 10 & 3-way pure epistasis & 3 &  1 & NA & 1 $\&$ 0.2 & 3200\\ \hline
11 $\&$ 12 & additive (2-way epistasis $+$ univariate) & 4 &  3 & equal & 1 $\&$ 0.4 & 1600\\ \hline
13 $\&$ 14 & additive (2, 2-way epistasis) & 4 & 2 & equal & 1 $\&$ 0.4 & 1600\\ \hline
15 $\&$ 16 & heterogeneity (univariate)  & 2 &  2 & equal & 1 $\&$ 0.4 & 1600\\ \hline
17 $\&$ 18 & heterogeneity (univariate) & 4 &  4 & equal & 1 $\&$ 0.4 & 1600\\ \hline
19 $\&$ 20 & heterogeneity (2, 2-way epistasis) & 4 & 2 & equal & 1 $\&$ 0.4 & 1600\\ \hline
21 & heterogeneity (2, 2-way epistasis) & 4 & 2 & $75:25$ & 0.4 & 1600\\ \hline
\end{tabular}
\end{table}

The simulation studies above are applied to map out FT signatures for different patterns of association in controlled scenarios where the 'ground truth' is known. Lastly, we applied LCS-DIVE to a real-world study of pancreatic cancer targeting 2 datasets comprised of survey variables that measure both established and potential risk factors for pancreatic cancer (Table \ref{tab:realparams}). The `P1' data was derived from the Prostate, Lung, Colorectal, and Ovarian Cancer (PLCO) Trial which began recruitment in 1993 and was previously detailed in \cite{prorok2000design}. The `P2' data adds dietary features from the initial baseline survey and a second self-administered dietary questionnaire that was distributed between 1998 and 2005 \cite{subar2000evaluation}. See Sup.2.3 for details on PLCO and target datasets. Here, these dataset are applied as an example of applying LCS-DIVE, and what has been learned from simulations, to real world data. 

\begin{table}[tph]
\centering
\captionof{table}{Pancreatic Cancer Dataset Summary} \label{tab:realparams}
\begin{tabular}{|c|c|c|c|c|}
\hline
Dataset & Cases & Controls & $\#$ Features & Feature Types \\
\hline
P1           & 800         & 4298 & 23 & Established Risk Factors and Covariates\\
\hline
P2             & 800         & 4298 & 37 & + Dietary features\\
\hline
\end{tabular}
\end{table}

\section{Results and Discussion}
This section presents results for applying LCS-DIVE across all datasets described in section \ref{data_sec}. Particular focus is paid to interpretation of FT signatures towards characterizing different underlying patterns of association, as well as identifying heterogeneous instance subgroups where applicable (i.e. Phase 2). LCS-DIVE generates a large number of figures and results to guide interpretation. Here we present key findings and visualizations and make further results available in Sup.3.

\subsection{Characterizing Epistatic and Heterogeneous Associations in Clean \emph{x}-MUX Problems}
\begin{figure}[tph]
    \centering
    \subfloat[\centering 6-bit MUX]{{\includegraphics[width=0.4\linewidth, keepaspectratio]{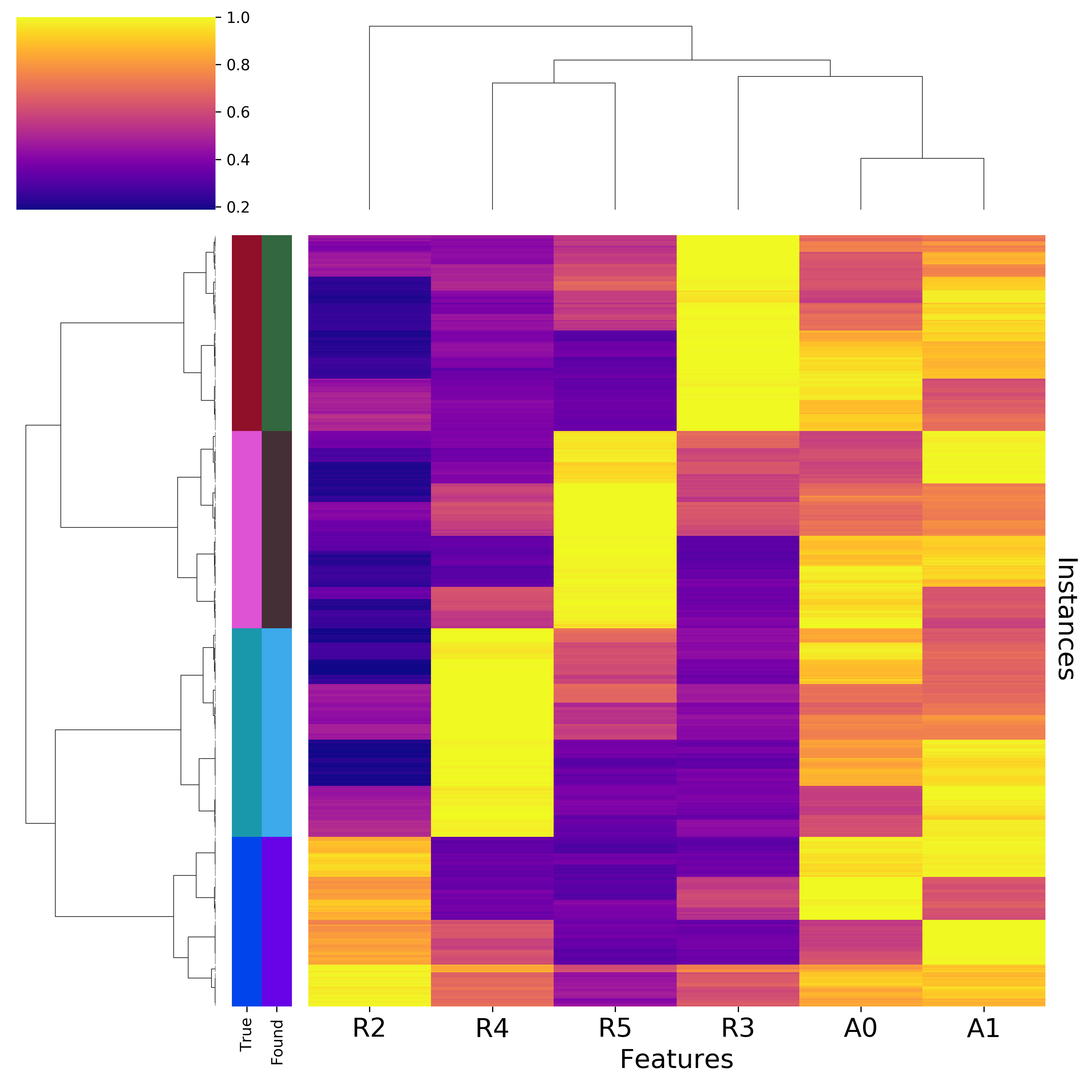}}}%
    \qquad
    \subfloat[\centering 20-bit MUX]{{\includegraphics[width=0.4\linewidth, keepaspectratio]{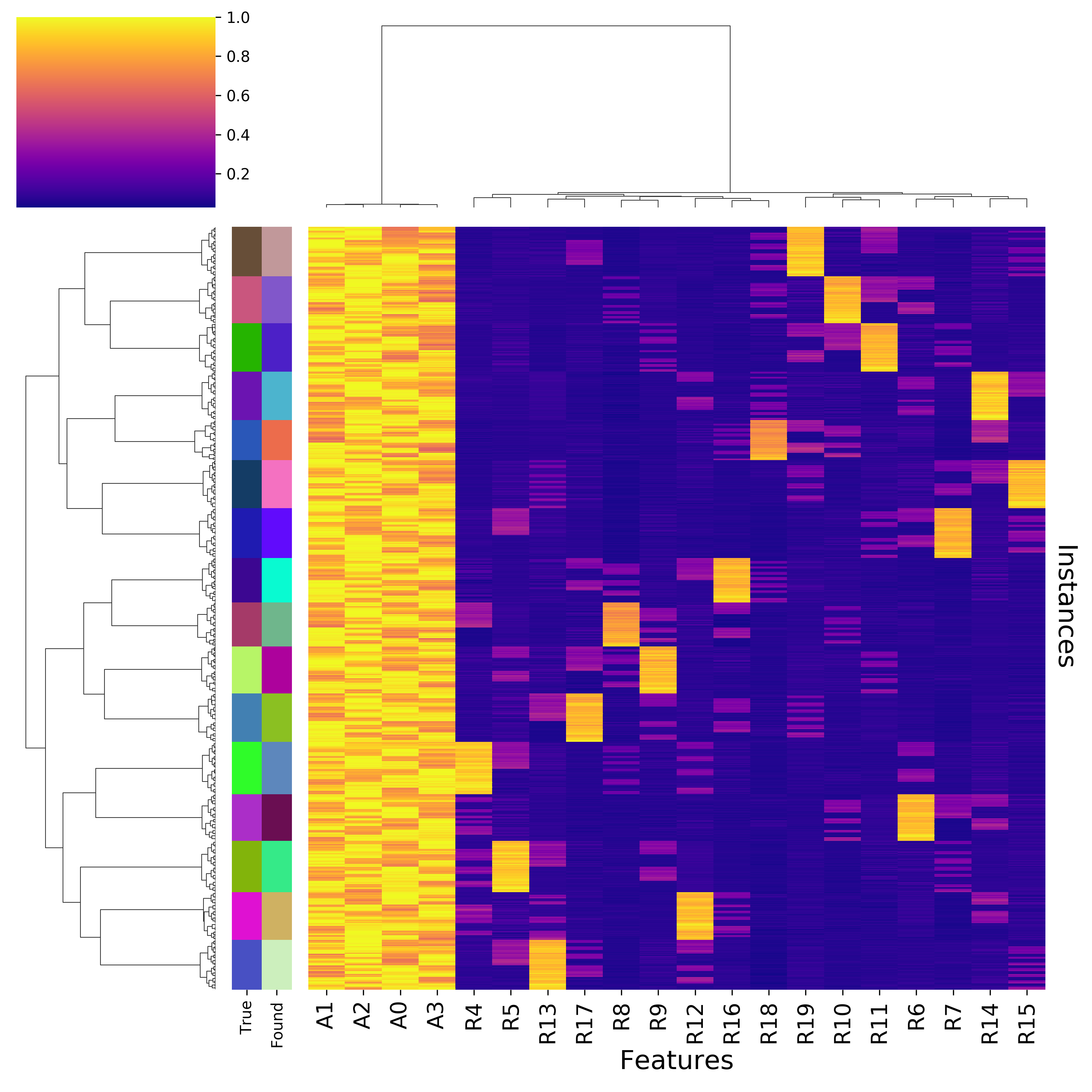}}}%
    \qquad
    \subfloat[\centering 37-bit MUX]{{\includegraphics[width=0.4\linewidth, keepaspectratio]{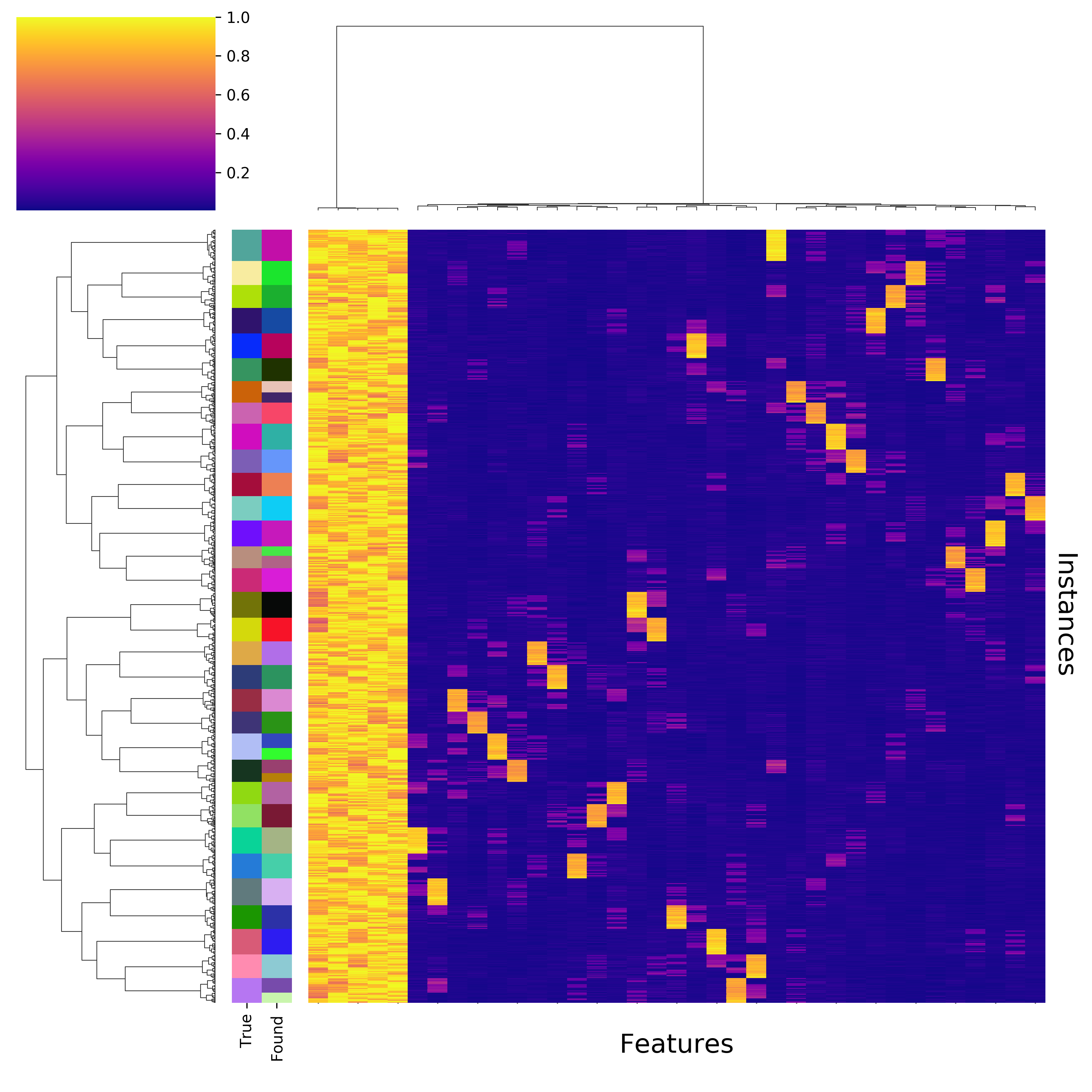}}}%
    \qquad
    \subfloat[\centering 70-bit MUX]{{\includegraphics[width=0.4\linewidth, keepaspectratio]{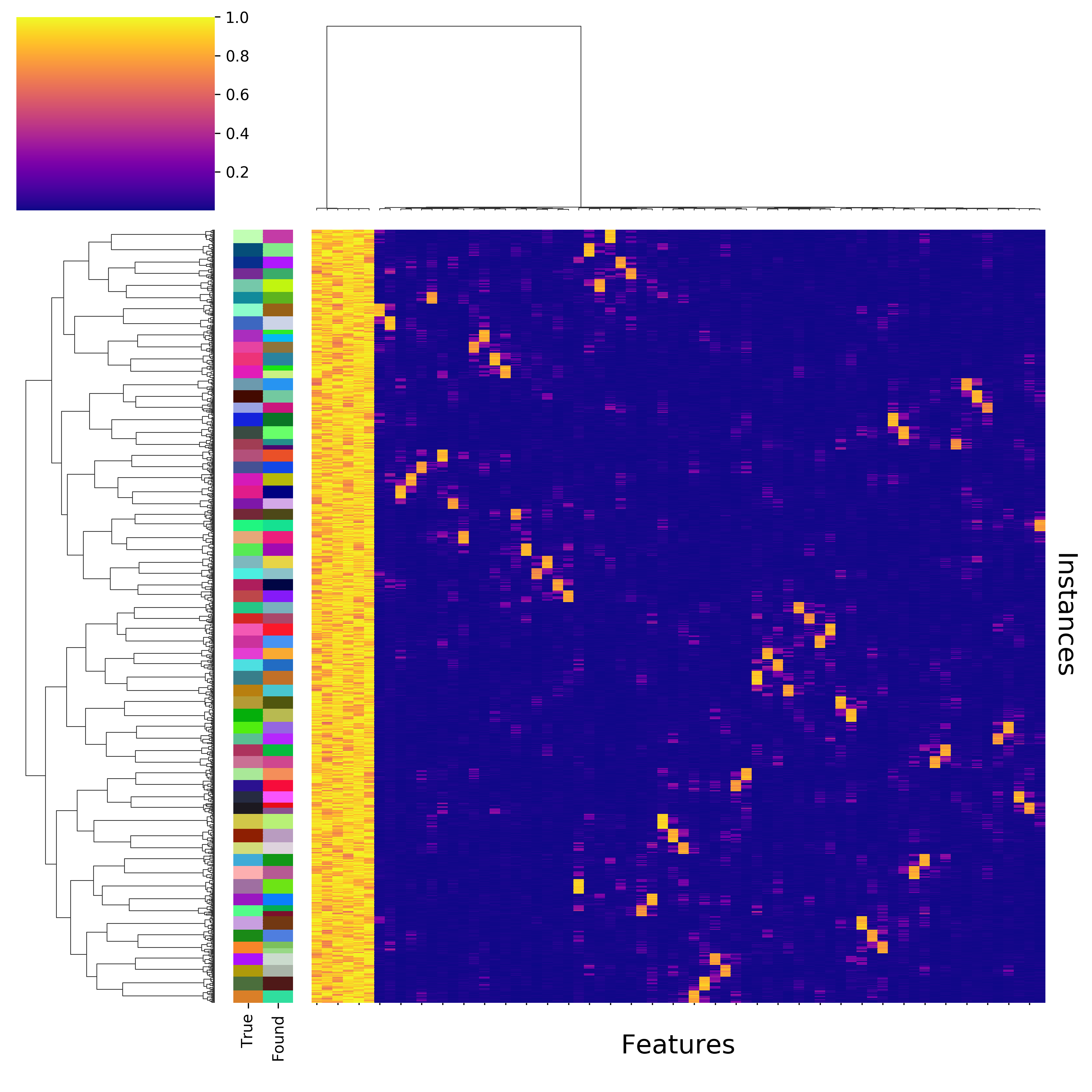}}}%
    \caption{FT Clustermaps of 6, 20, 37, and 70-bit MUX datasets. For space we exclude feature labels on the 37 and 70-bit clustermaps, but address bit features are the 5 or 6 leftmost columns, respectively (with a consistent yellow signature)}
    \label{fig:muxplots}
\end{figure}

LCS-DIVE was first applied to the MUX datasets (see Table \ref{tab:muxparams}). Phase 1 modeling with scikit-ExSTraCS yielded $100\%$ average testing accuracy for all 5 datasets. This is consistent with previous ExSTraCS 2.0 analyses \cite{urbanowicz2015exstracs} supporting successful implementation of scikit-ExSTraCS. Figure \ref{fig:muxplots} gives clustermaps for remaining MUX datasets, with the 11-bit MUX previously illustrated in Figure \ref{fig:ATSample}. For each \emph{x}-MUX, the automated elbow approach was applied and successfully picked the optimal number of clusters as defined by the 'heterogeneous combinations' column of Table \ref{tab:muxparams} (see Sup.3.1). We confirmed that `true' clusters always aligned with `found' clusters as seen in Figure \ref{fig:muxplots}. One exception is noted for the 70-bit MUX: 65 clusters were proposed instead of the optimal 64, having split one of the true clusters in half. 

The underlying epistatic interaction within each subgroup is illustrated by the combination of features yielding higher FT scores within the respective subgroup (e.g. For the 37-bit MUX: 6-way interactions, i.e. 5 address and 1 register bit). For all MUX problems, we expect for any row (i.e. instance), all address bits and a single register bit should yield highest FT scores. This is largely the case but with some observable variation (particularly in the 6-bit MUX). This variation is the result of: (1) FT scoring during modeling can be impacted by sub-optimal rules that continuously appear and are eliminated during the evolutionary search, and (2) MUX problems can also be solved with the discovery of rules that are not globally optimal, but still $100\%$ accurate, previously described as being `natural' rather than `optimal' solutions \cite{liu2019absumption}. Despite these imperfections, Phase 2 LCS-DIVE successfully and automatically detected all heterogeneous subgroups and corresponding epistatic interactions for each MUX problem. An example of Phase 3 rule population visualizations for the 6-bit MUX is available in Sup.3.1.1.

\subsection{Characterizing Distinct Patterns of Association in Simulated GAMETES Datasets}
This section applies LCS-DIVE to the simulated SNP datasets described in Table \ref{tab:simdata}. Phase 1 modeling with scikit-ExSTraCS yielded expected average testing accuracies for the 21 datasets demonstrating successful modeling of the underlying associations. Sup.3.2 details these accuracies along with the automatically suggested number of `found' clusters for each. We observe that this approach consistently overestimates the number of clusters, particularly in the context of homogeneous associations, or in datasets with noise. Thus the elbow plot mechanism may be used as a guide for interpretation, but ultimately users should also rely on subjective examination of the 1 to \emph{k} proposed cluster outputs by LCS-DIVE. Below, we take a closer look at the FT clustermaps generated for the different data scenarios and identify guidelines for interpreting unique FT signatures. In all figures, features starting with `M' (e.g. M0P1) are predictive features and those starting with `N' are randomly generated (i.e. uninformative) as describe in \cite{urbanowicz2012gametes}.

\subsubsection{GAMETES: Simple Univariate and Additive Associations}
\begin{figure}[tph]
    \centering
    \subfloat[\centering D1: Clean Univariate]{{\includegraphics[width=0.4\linewidth, keepaspectratio]{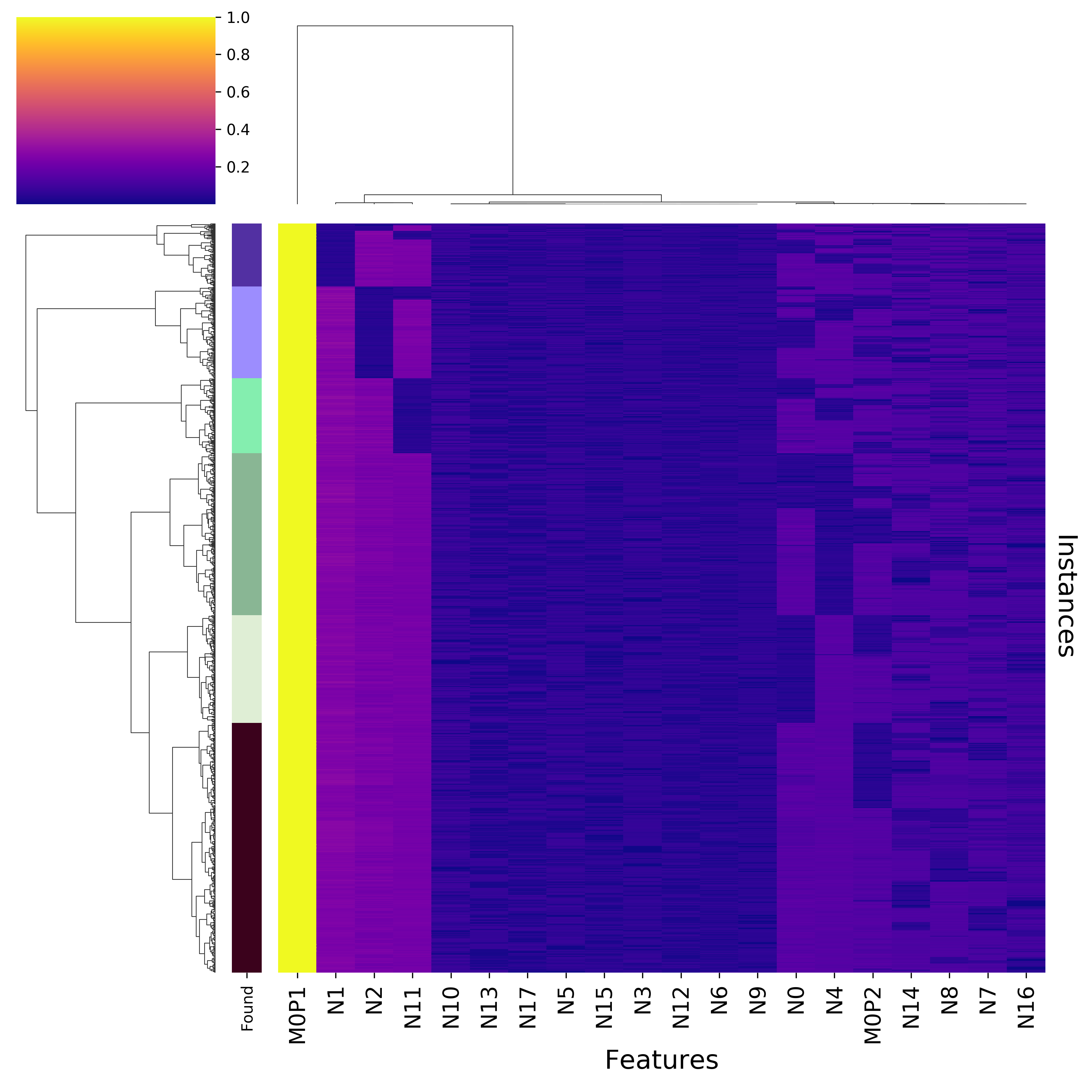}}}%
    \qquad
    \subfloat[\centering D2: Noisy Univariate]{{\includegraphics[width=0.4\linewidth, keepaspectratio]{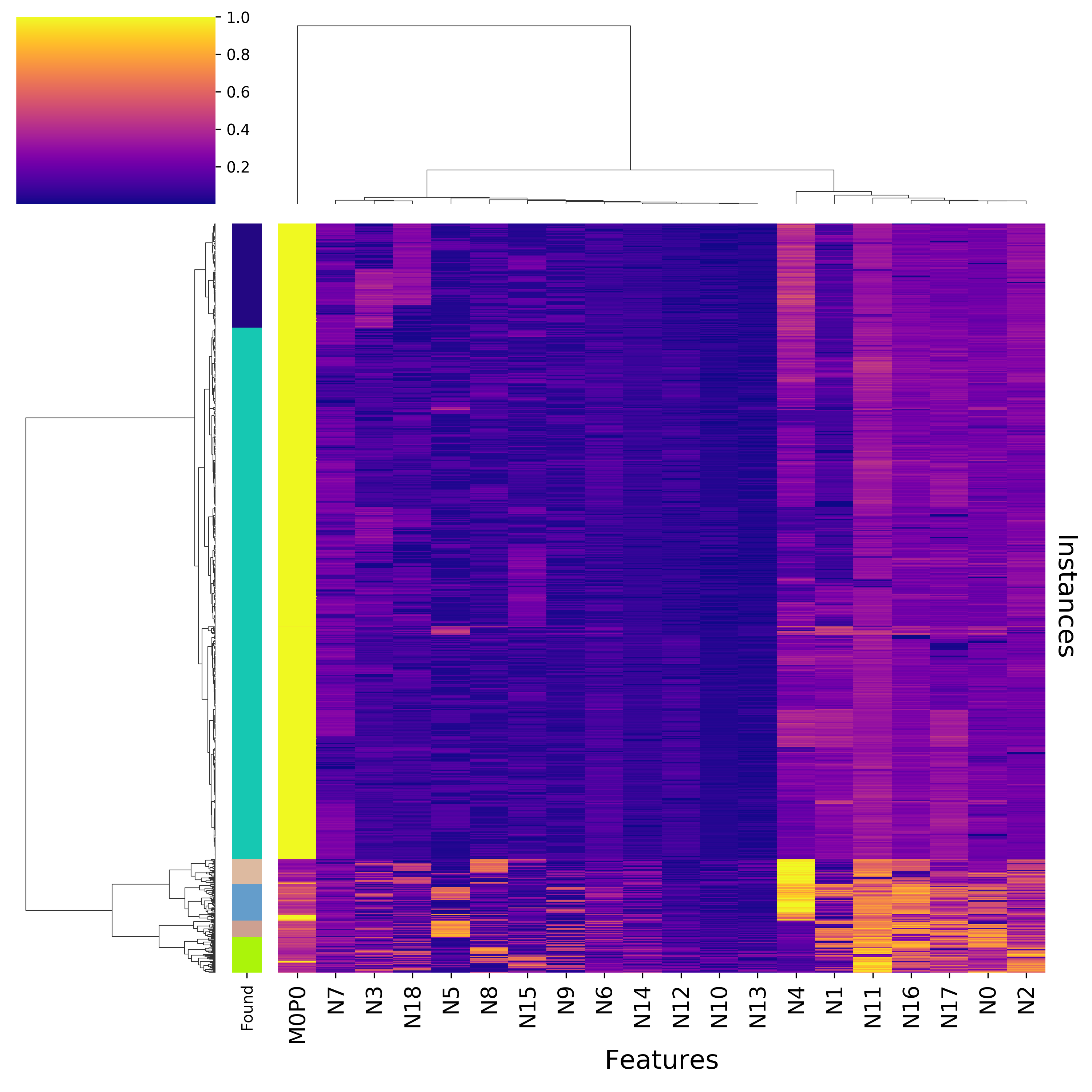}}}%
    \qquad
    \subfloat[\centering D3: Clean 2-feature additive]{{\includegraphics[width=0.4\linewidth, keepaspectratio]{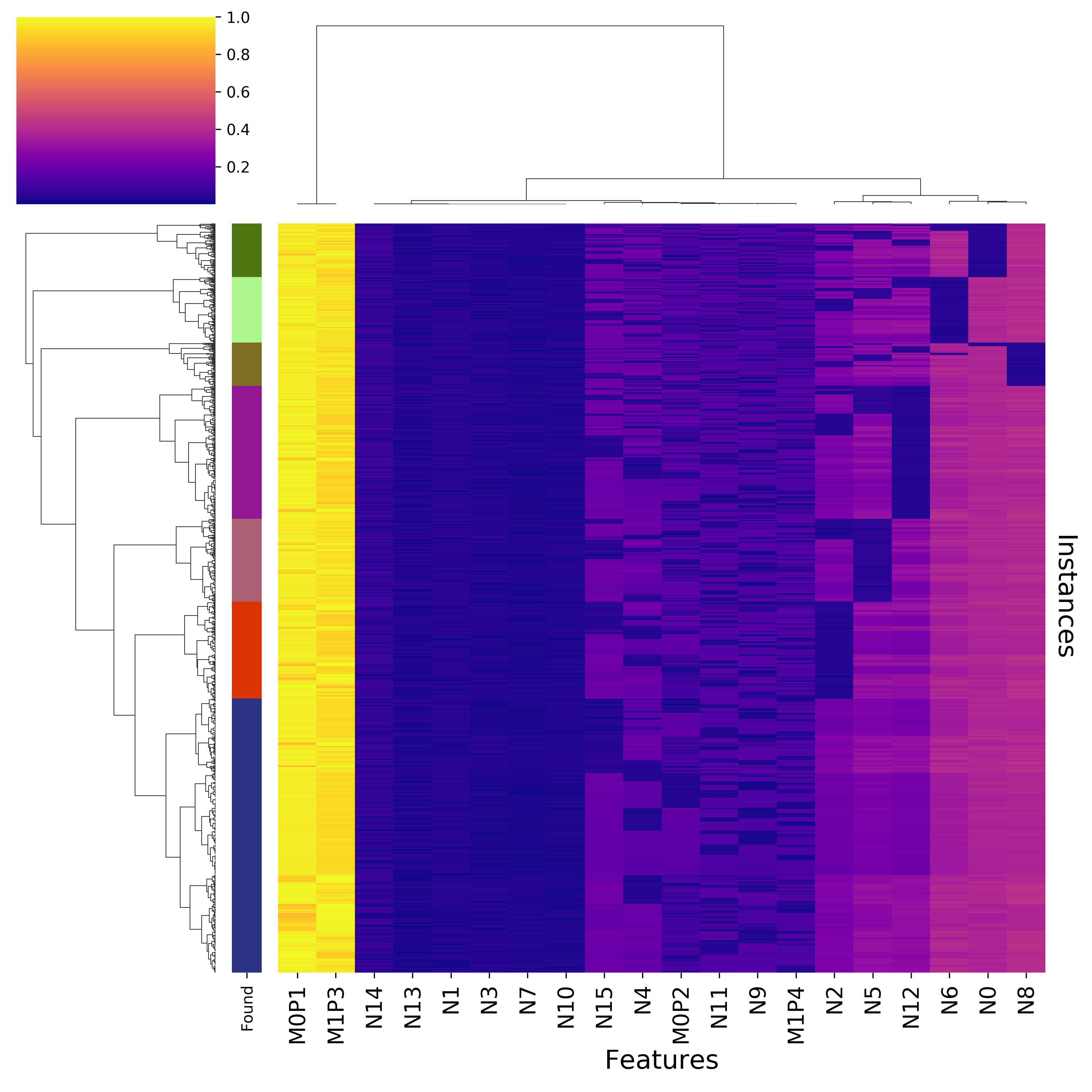}}}%
    \qquad
    \subfloat[\centering  D4: Noisy 2-feature additive]{{\includegraphics[width=0.4\linewidth, keepaspectratio]{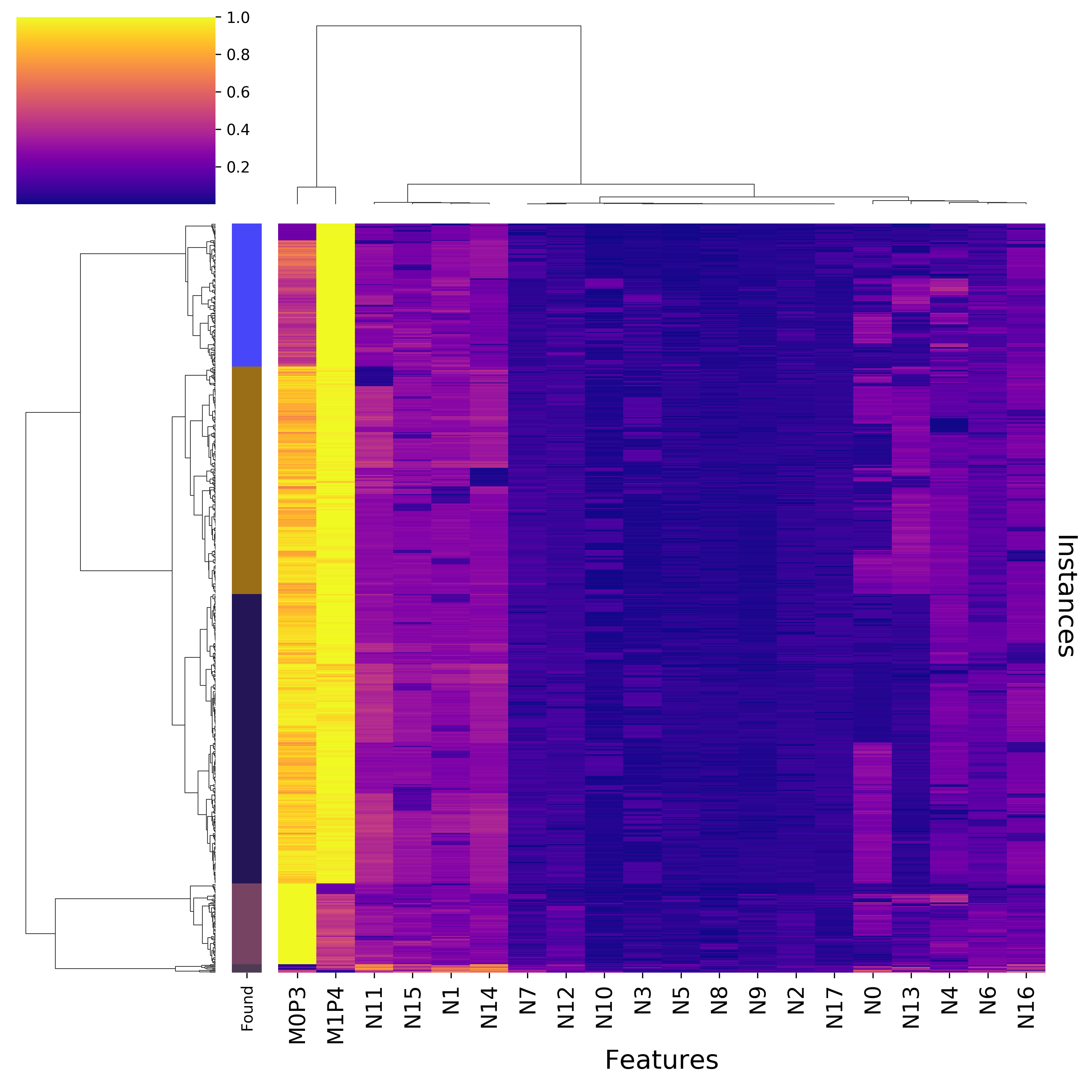}}}%
    \qquad
    \subfloat[\centering D5: Clean 4-feature additive]{{\includegraphics[width=0.4\linewidth, keepaspectratio]{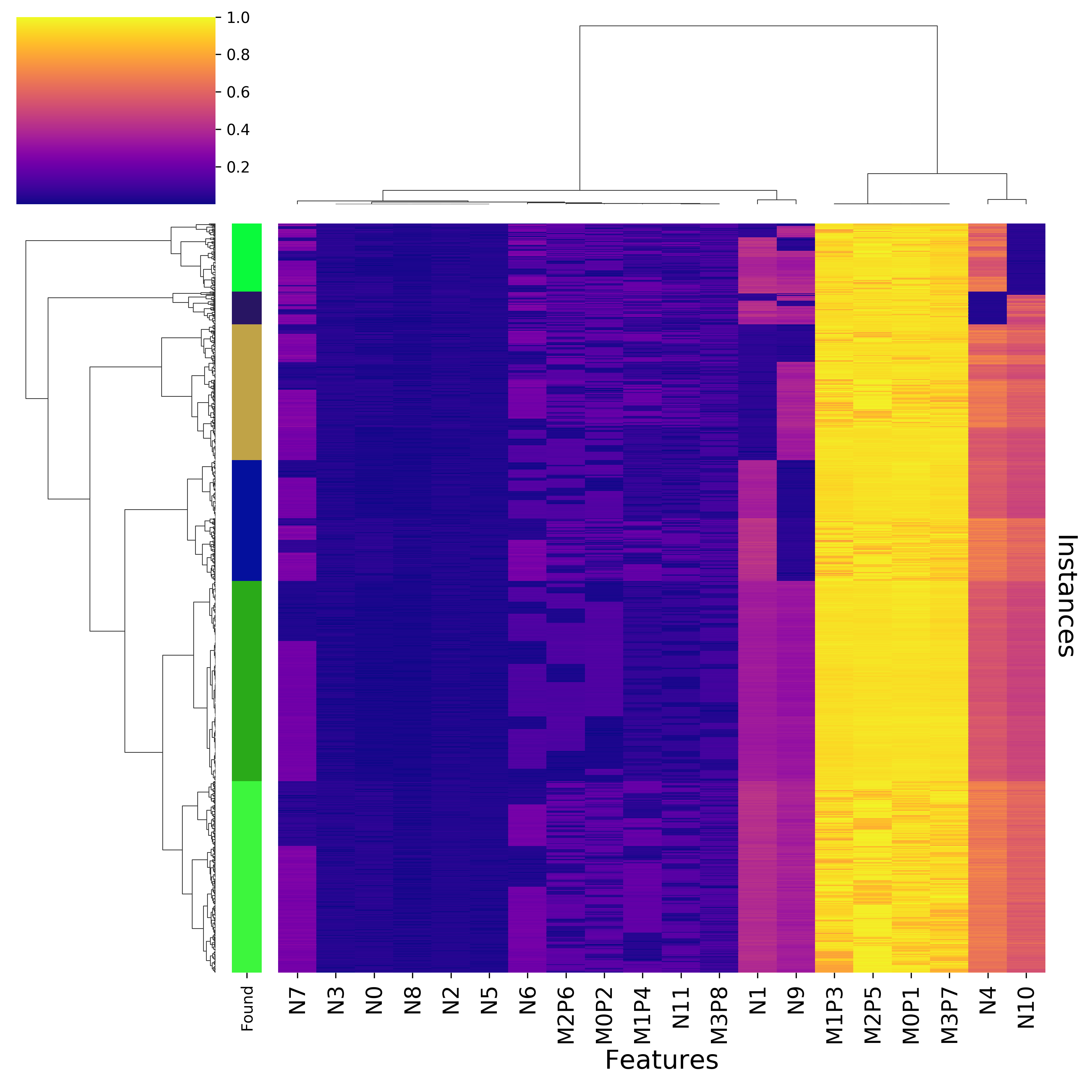}}}%
    \qquad
    \subfloat[\centering D6: Noisy 4-feature additive]{{\includegraphics[width=0.4\linewidth, keepaspectratio]{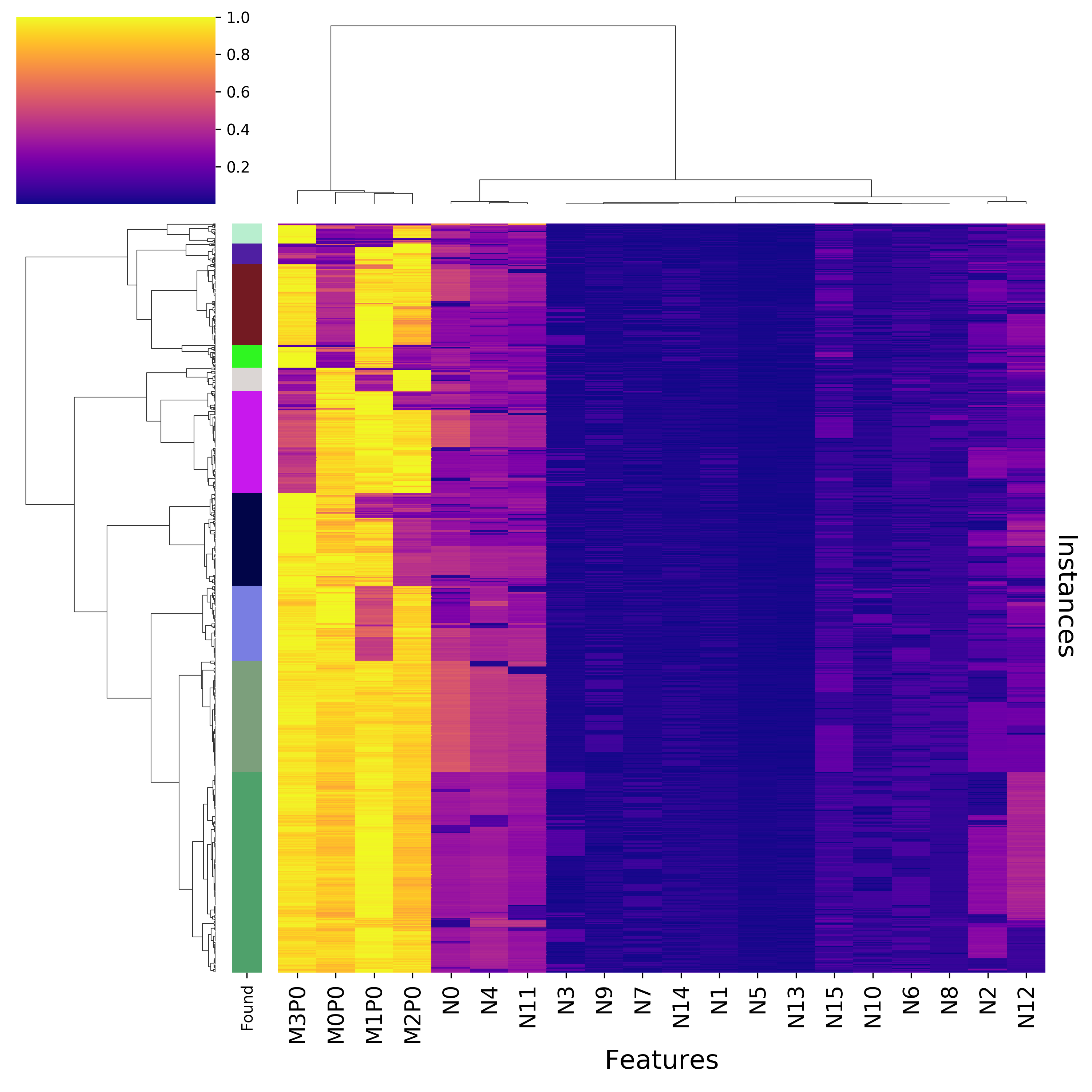}}}%
    \caption{FT Clustermaps of univariate or additive GAMETES datasets D1-D6}
    \label{fig:ME}
\end{figure}

Here we examine clean and noisy GAMETES datasets D1-D6 with either a univariate, 2-feature additive, or 4-feature additive association. FT clustermaps for each of the 6 datasets are given in Figure \ref{fig:ME}. As no heterogeneity is simulated here, no `true' subgroups are identified.

For D1 (Figure \ref{fig:ME}A), a strong signature for the single predictive feature is observed for all instances. Notably the 'found' clusters are incorrectly picking up on subtle FT signatures in some of the non-predictive features, causing more than one 'optimal' cluster to be identified. An example of Phase 3 rule population visualizations for D1 is available in Sup.3.2.1. For D2, (Figure \ref{fig:ME}B) there is a similar univariate signature however a proportion of instances in lower clusters yield a messy signature. These correspond to noisy instances that didn't follow the modeled univariate pattern. We know from this controlled simulation that these lower clusters are not representative of underlying heterogeneity. LCS-DIVE automatically exports testing accuracy for the instances within respective clusters. From this we confirm that testing accuracy is very high in the top two clusters (capturing the univariate association) but close to zero in the lower `noisy' clusters. Later, this observation will allow us to differentiate between clusters that represent candidate heterogeneous subgroups, and clusters that only capture underlying noise.

For D3 (Figure \ref{fig:ME}C), we observe a largely strong signature for both features. Notably, this signature is less sharp compared to D1 because: either or both of the predictive features can be applied within a given instance to correctly predict class, but both are not required. This yields LCS rules that specify just one or both features, impacting the resulting FT signature. For D4, (Figure \ref{fig:ME}D), we can observe 4 unique clusters, i.e. two clusters where only one feature is predictive (because there is noise in the other), a cluster where both features are predictive on their own, and a small cluster at the bottom corresponding to completely noisy instances. 

Similarly, for D5 (Figure \ref{fig:ME}E) we observe the strongest signature for the 4 additive predictive features together, however for D6 (Figure \ref{fig:ME}F) we observe one larger cluster of instances where all of the features appear predictive, and a number of smaller clusters capturing instances where some subset of the four features was predictive but one or more of the four was not. This pattern of clustering defines a unique signature for additive univariate effects in the context of FT clustering that other patterns of association do not display (shown below). 

\subsubsection{GAMETES: Epistatic Associations}
\begin{figure}[tph]
    \centering
    \subfloat[\centering D7: Clean 2-way pure epistasis]{{\includegraphics[width=0.4\linewidth, keepaspectratio]{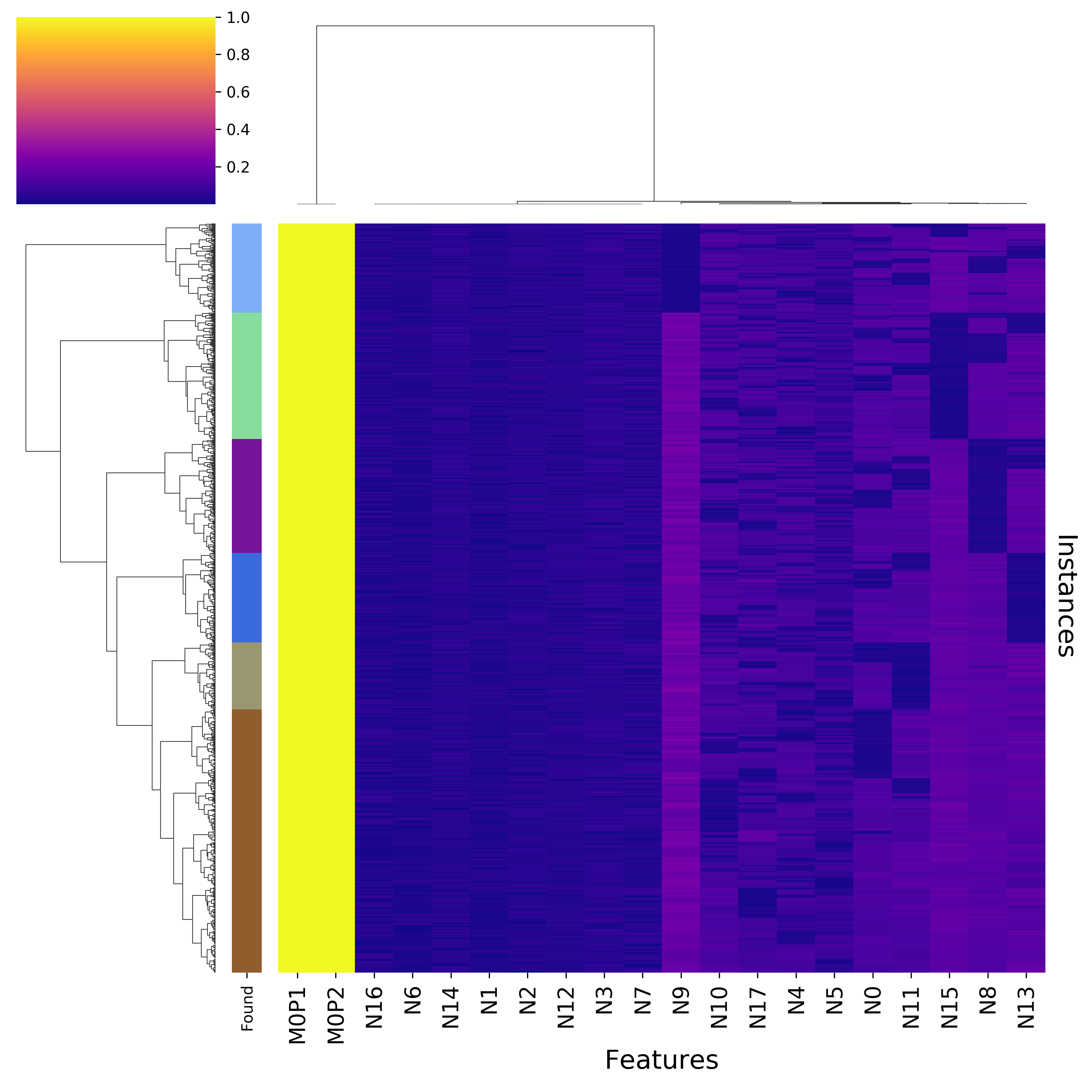}}}%
    \qquad
    \subfloat[\centering D8: Noisy 2-way pure epistasis]{{\includegraphics[width=0.4\linewidth, keepaspectratio]{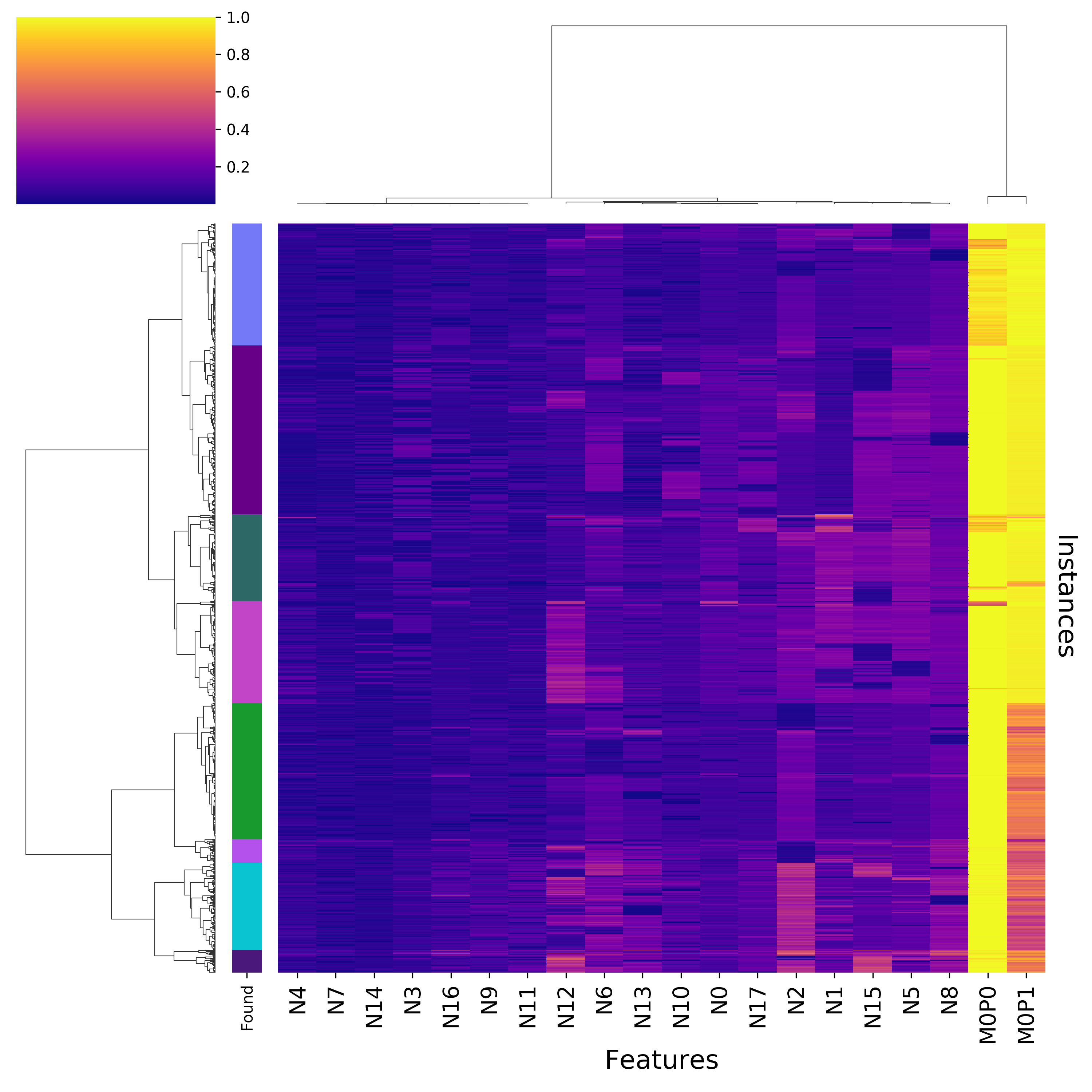}}}%
    \qquad
    \subfloat[\centering D9: Clean 3-way pure pistasis]{{\includegraphics[width=0.4\linewidth, keepaspectratio]{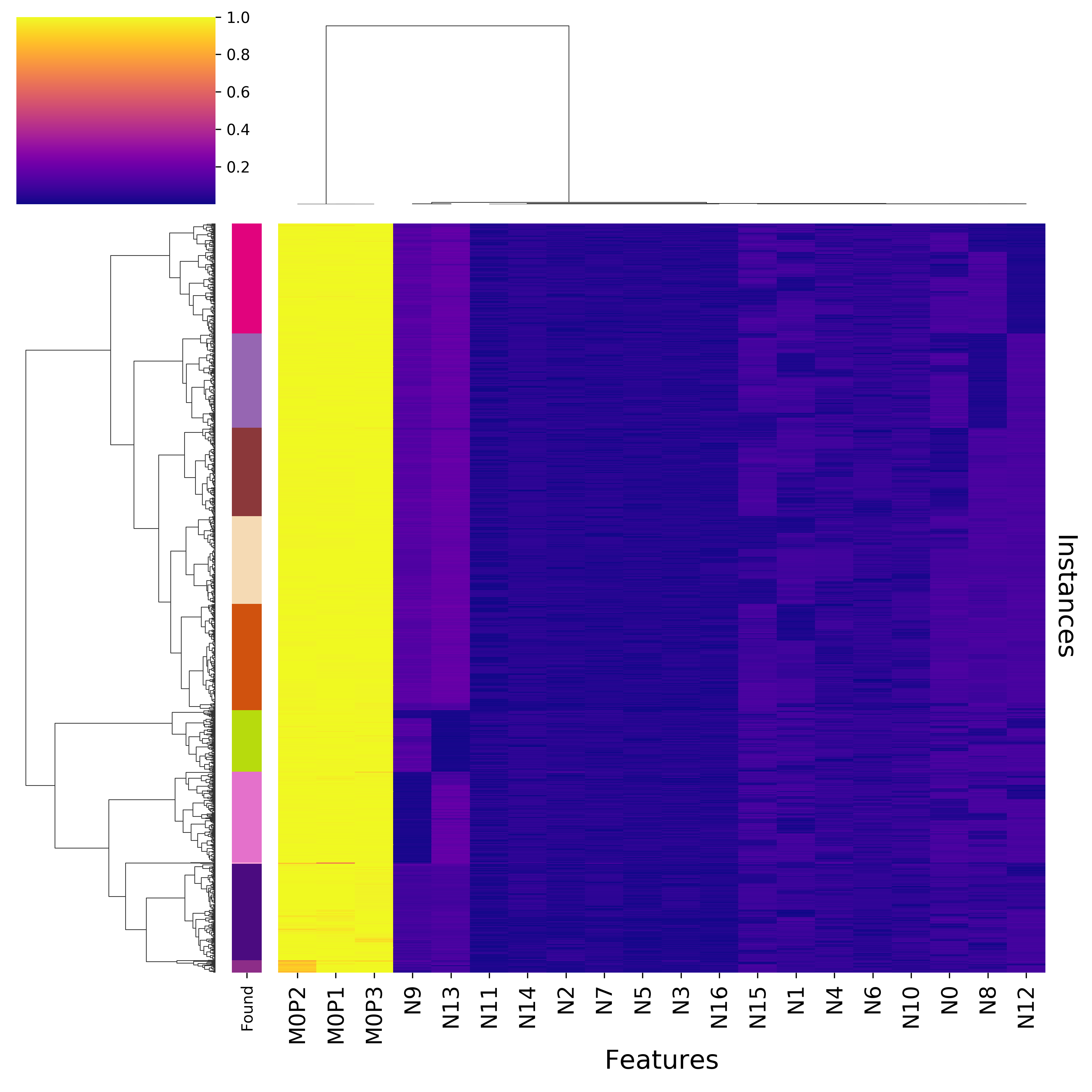}}}%
    \qquad
    \subfloat[\centering D10: Noisy 3-way pure epistasis]{{\includegraphics[width=0.4\linewidth, keepaspectratio]{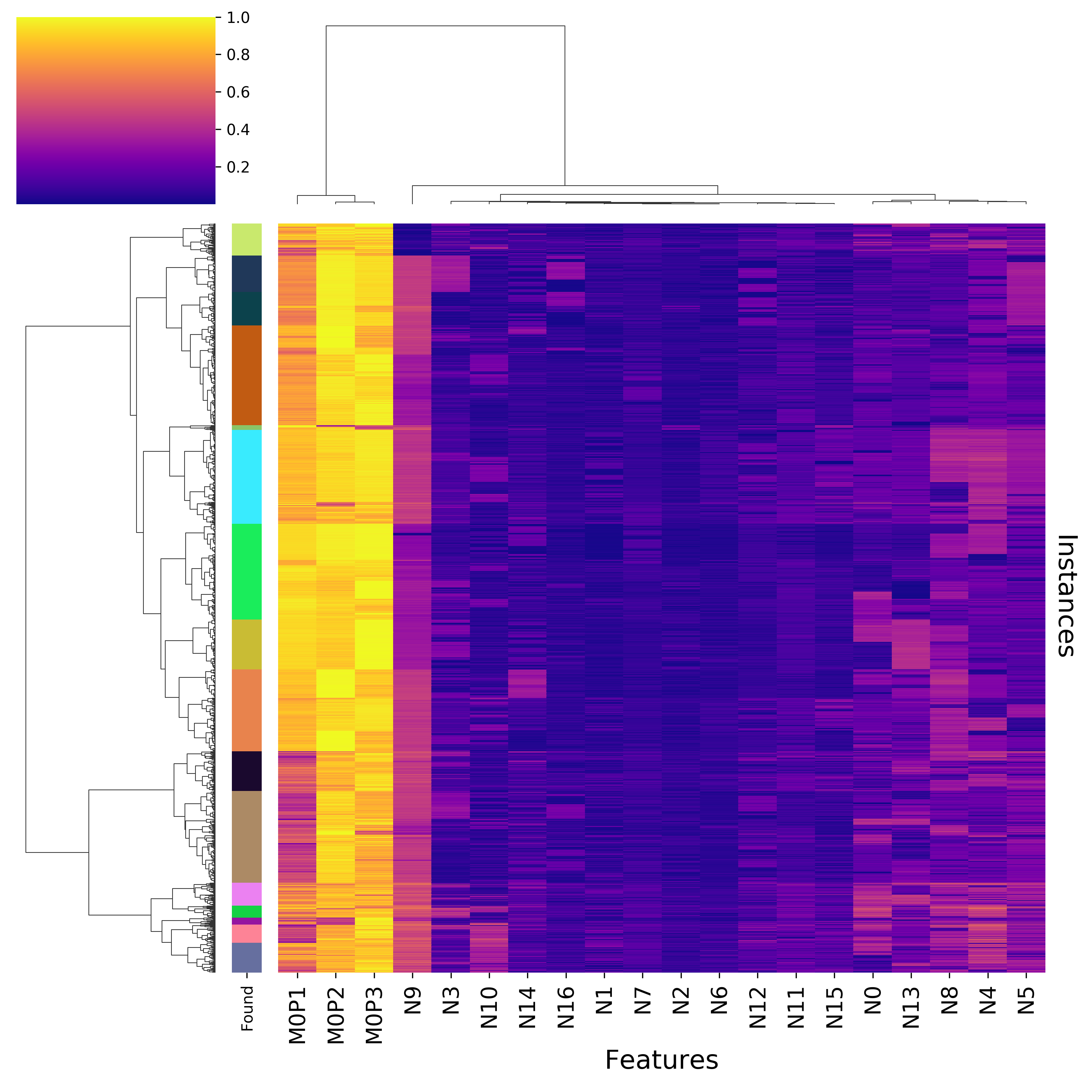}}}%
    \caption{FT Clustermaps of purely epistatic GAMETES datasets D7-D10}
    \label{fig:Epi1}
\end{figure}

\begin{figure}[tph]
    \centering
    \subfloat[\centering D11:Clean additive (2-way epistasis+univariate)]{{\includegraphics[width=0.4\linewidth, keepaspectratio]{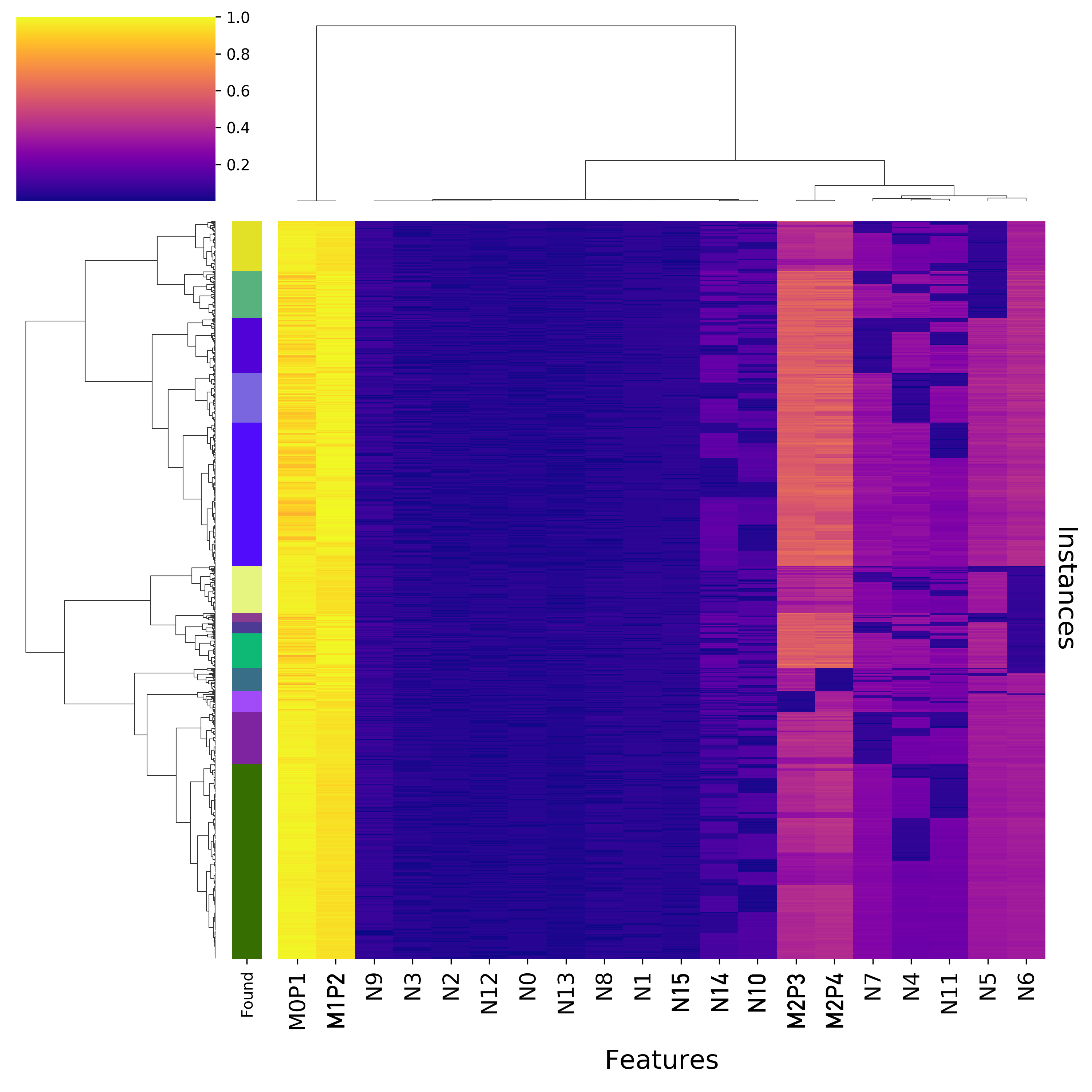}}}%
    \qquad
    \subfloat[\centering D12: Noisy additive (2-way epistasis+univariate)]{{\includegraphics[width=0.4\linewidth, keepaspectratio]{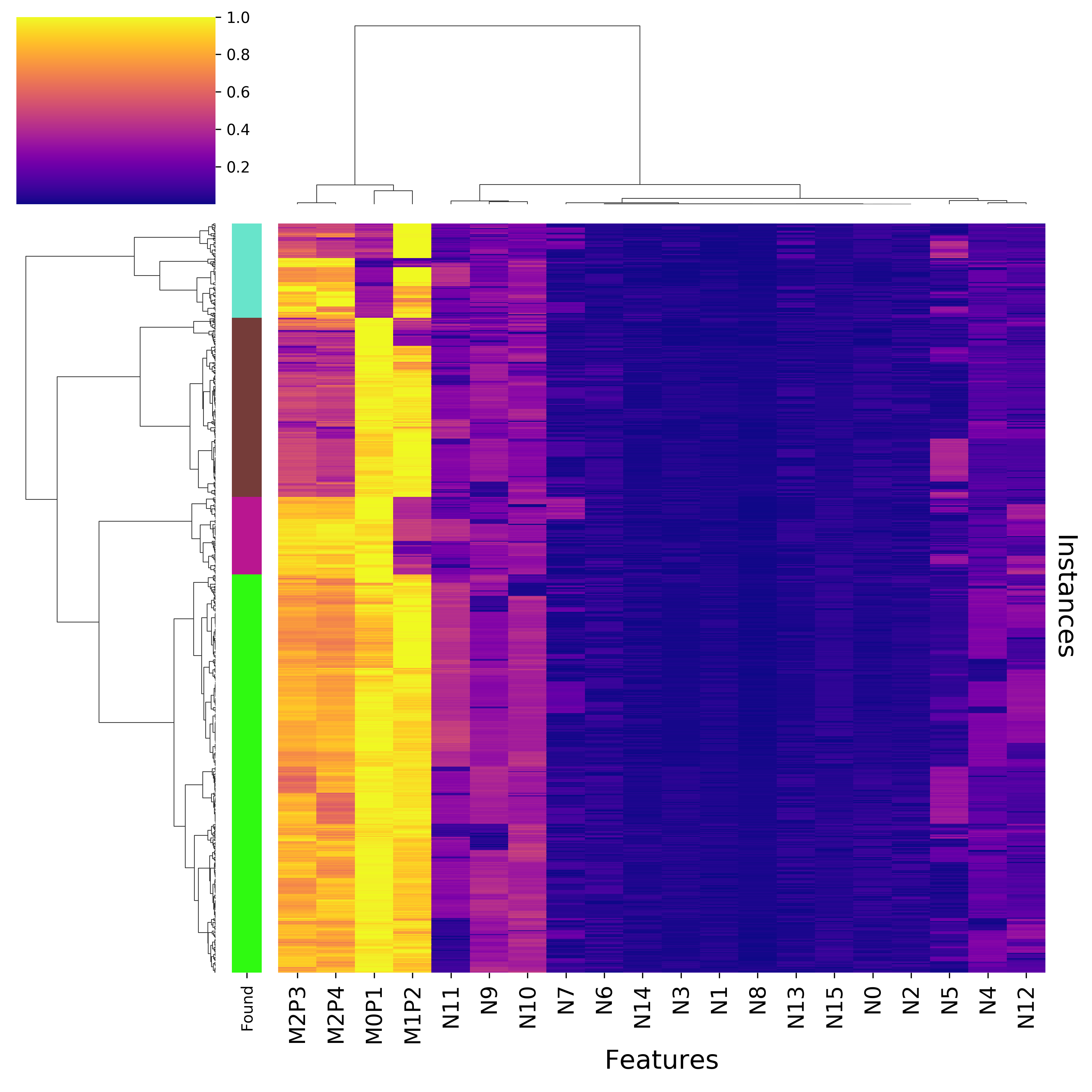}}}%
    \qquad
    \subfloat[\centering D13: Clean additive (2, 2-way epistasis) ]{{\includegraphics[width=0.4\linewidth, keepaspectratio]{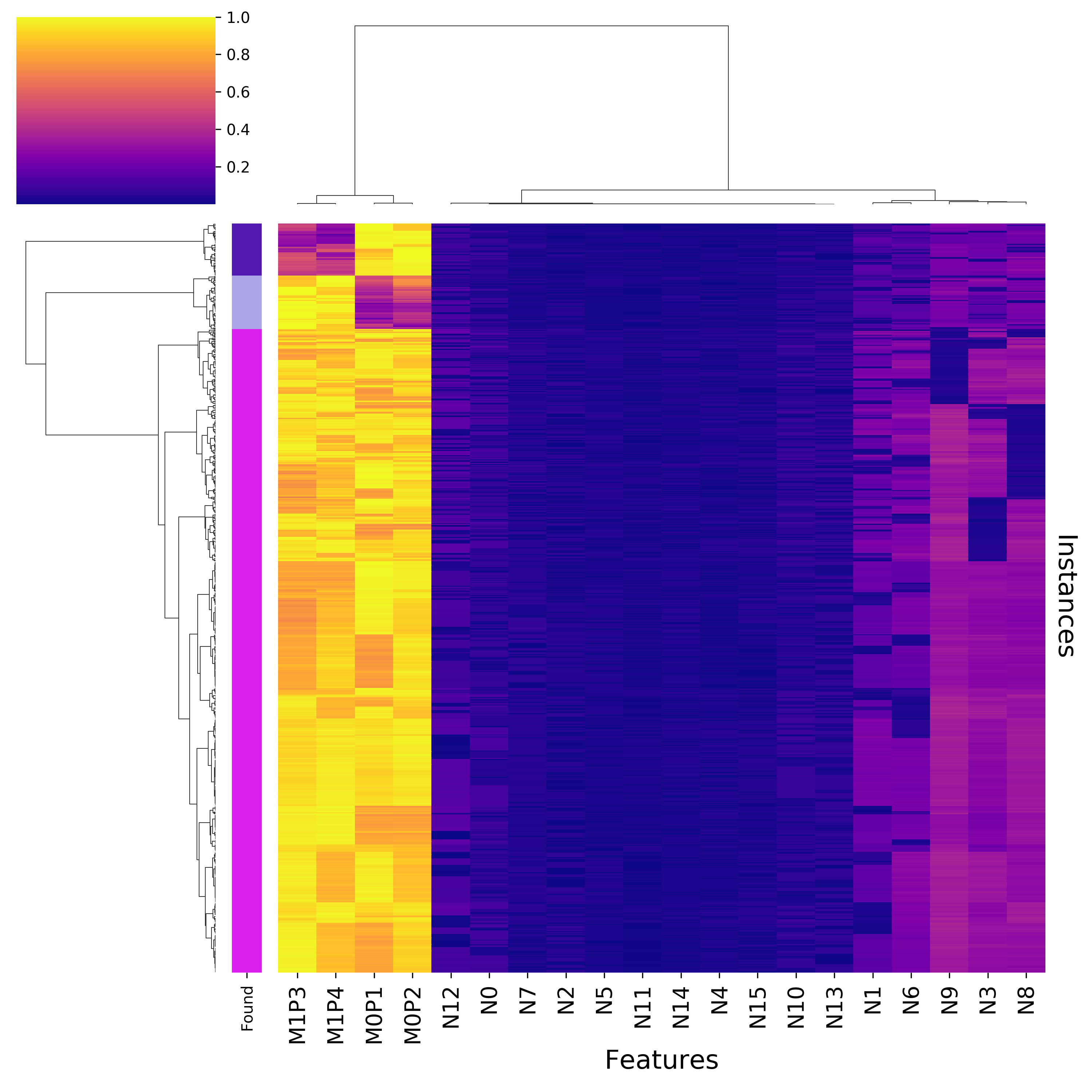}}}%
    \qquad
    \subfloat[\centering D14: Noisy additive (2, 2-way epistasis)]{{\includegraphics[width=0.4\linewidth, keepaspectratio]{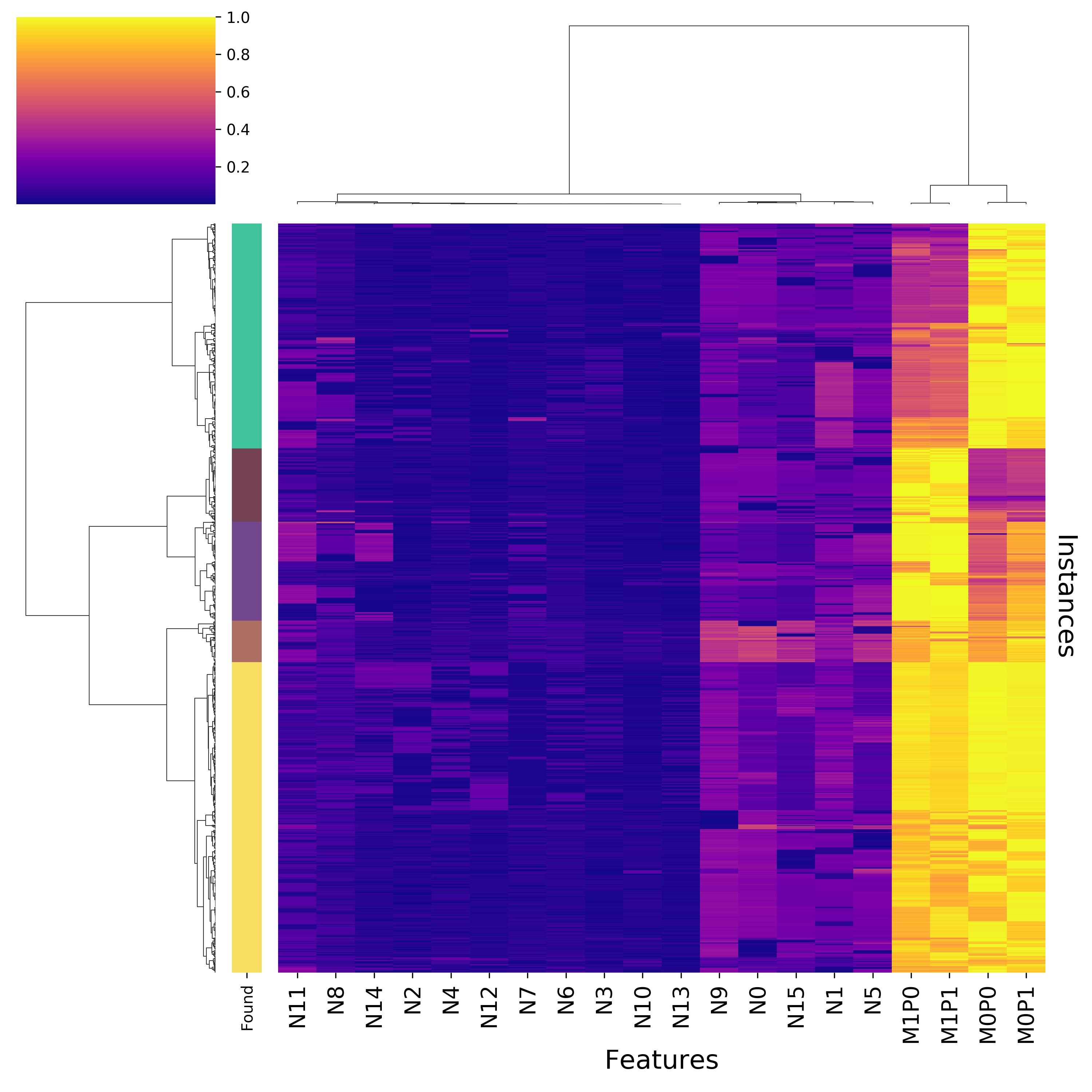}}}%
    \caption{FT Clustermaps of additively combined epistatic GAMETES datasets D11-D14}
    \label{fig:Epi2}
\end{figure}

Here we examine clean and noisy GAMETES datasets D7-D14 examining scenarios of pure epistasis and additive epistasis combinations (i.e. impure epistasis \cite{urbanowicz2012gametes}). FT clustermaps for pure epistasis datasets are given in Figure \ref{fig:Epi1} and those for impure epistasis are given in Figure \ref{fig:Epi2}. 

For D7 (Figure \ref{fig:Epi1}A), we observe a strong signature for both interacting predictive features across all instances. Note the subtle difference between this and 2 clean additively combined features from Figure \ref{fig:ME}C. For D8 (Figure \ref{fig:Epi1}B), we observe a strong but less consistent signature for the 2 interacting features. Distinct from the noisy 2-feature additive example (Figure \ref{fig:ME}D), we don't see instance clusters where either feature has a strong signature on it's own. Since this distinction is subtle, we recommend users conduct explicit statistical tests confirming the presence of epistasis vs. additive effects in practice. Similarly, for D9 (Figure \ref{fig:Epi1}C), we observe a strong signature for the three interacting predictive features across all instances. For D10 (Figure \ref{fig:Epi1}D), we observe a reasonably strong but less consistent signature for the 3 interacting features. However we do not see the same `blocky' signature observed in noisy additive data (Figure \ref{fig:ME}F) where some features are predictive but not all.

Next we examine impure epistasis scenarios in which features M2P3 and M2P4 are epistatically interacting, and M0P1 and M1P2 are univariate. For D11 (Figure \ref{fig:Epi2}A) we observe a much stronger signature for the two univariate features and a much weaker one for the epistatic features. We believe this is because, LCS rule discovery is pressured to find the simplest rules possible (i.e. those that specify the fewest features). Given that accurate predictions can be obtained from either univariate feature alone, it is not surprising that the signature of the interacting features is much weaker. For D12 (Figure \ref{fig:Epi2}B), we observe signatures reminiscent of additive univariate associations (i.e. instance clusters where one univariate feature has low scores) as well as epistatic interactions (i.e. where both interacting features have a very strong signature in specific clusters). For D13 (Figure \ref{fig:Epi2}C), we observe a largely strong cluster where all four features are consistently high but but neither epistatic pair has as strong a signal as in the 2-way pure epistasis (Figure \ref{fig:Epi1}A). Further we observe two small instance clusters where only one of the to pairs has a particularly strong signal. We believe this to be an artifact of rule-based learning where the instances of one epistatic pair constitutes a rarer genotype combination than the other and thus cumulatively receive lower FT updates than the more common pair. For D14 (Figure \ref{fig:Epi2}D), we observe a similar, but more drawn out pattern than in D13 as a reflection of the underlying noise.

\subsubsection{GAMETES: Heterogeneous Associations}
\begin{figure}[tph]
    \centering
    \subfloat[\centering D15: Clean 2-feature heterogeneous univariate]{{\includegraphics[width=0.4\linewidth, keepaspectratio]{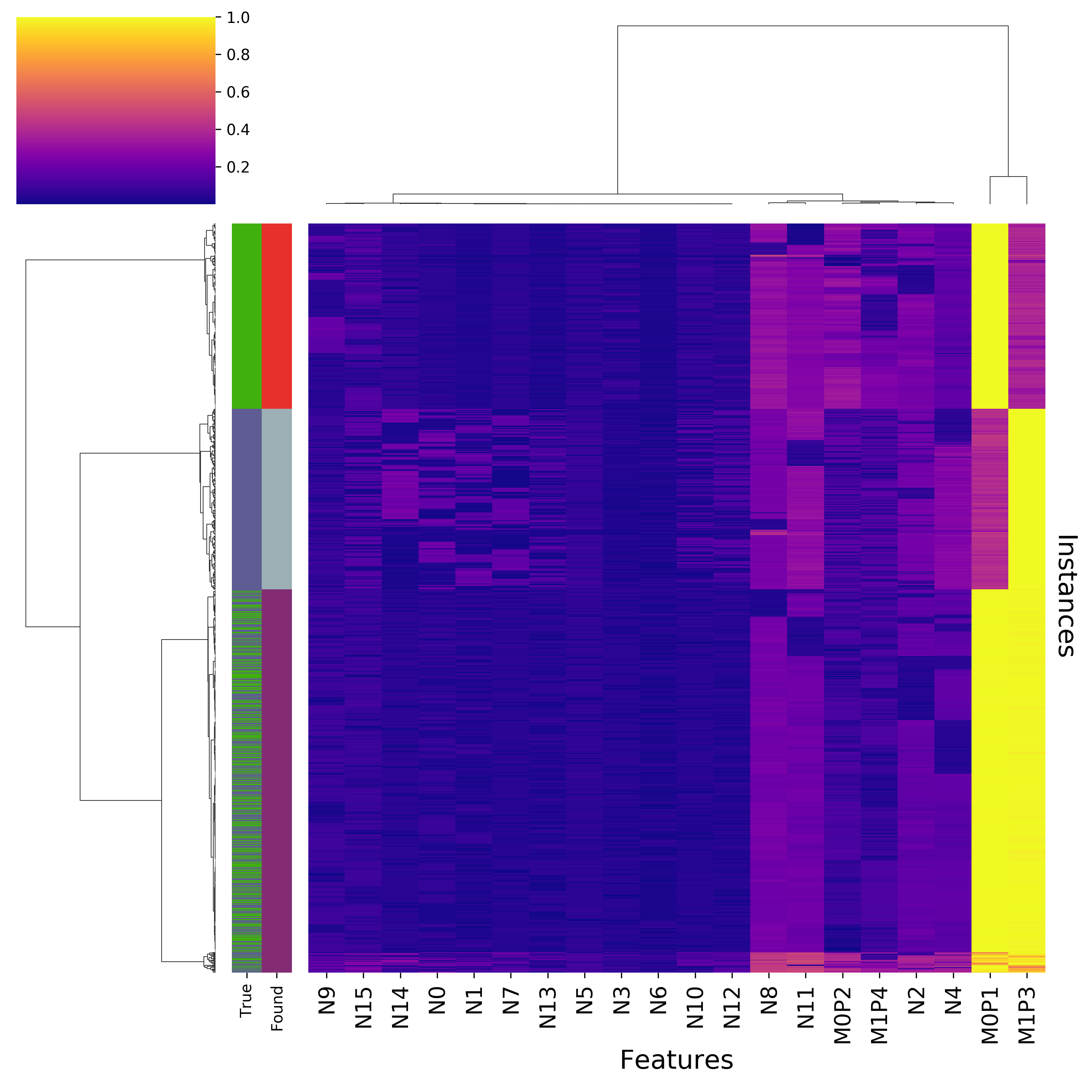}}}%
    \qquad
    \subfloat[\centering D16: Noisy 2-feature heterogeneous univariate]{{\includegraphics[width=0.4\linewidth, keepaspectratio]{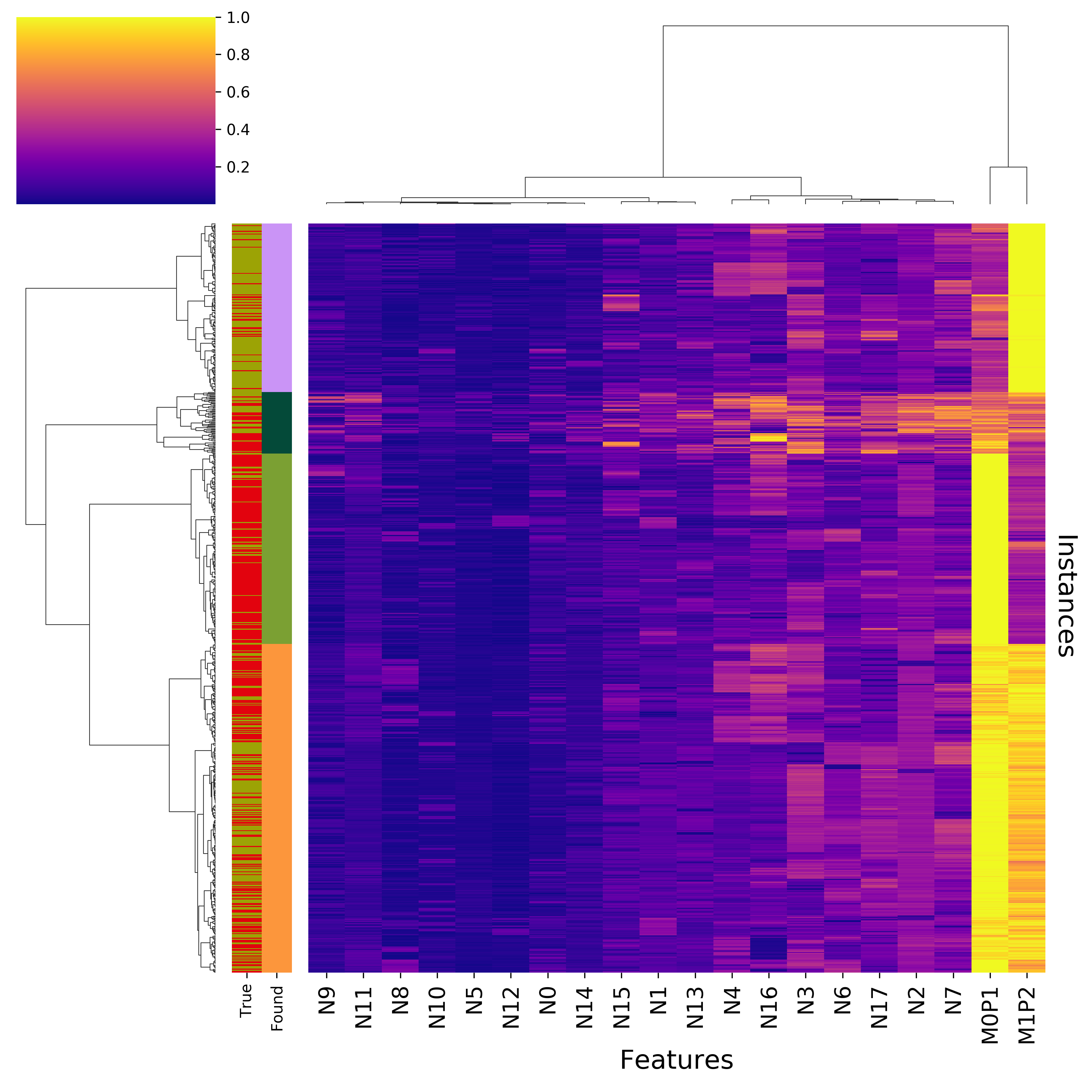}}}%
    \qquad
    \subfloat[\centering D17: Clean 4-feature heterogeneous univariate]{{\includegraphics[width=0.4\linewidth, keepaspectratio]{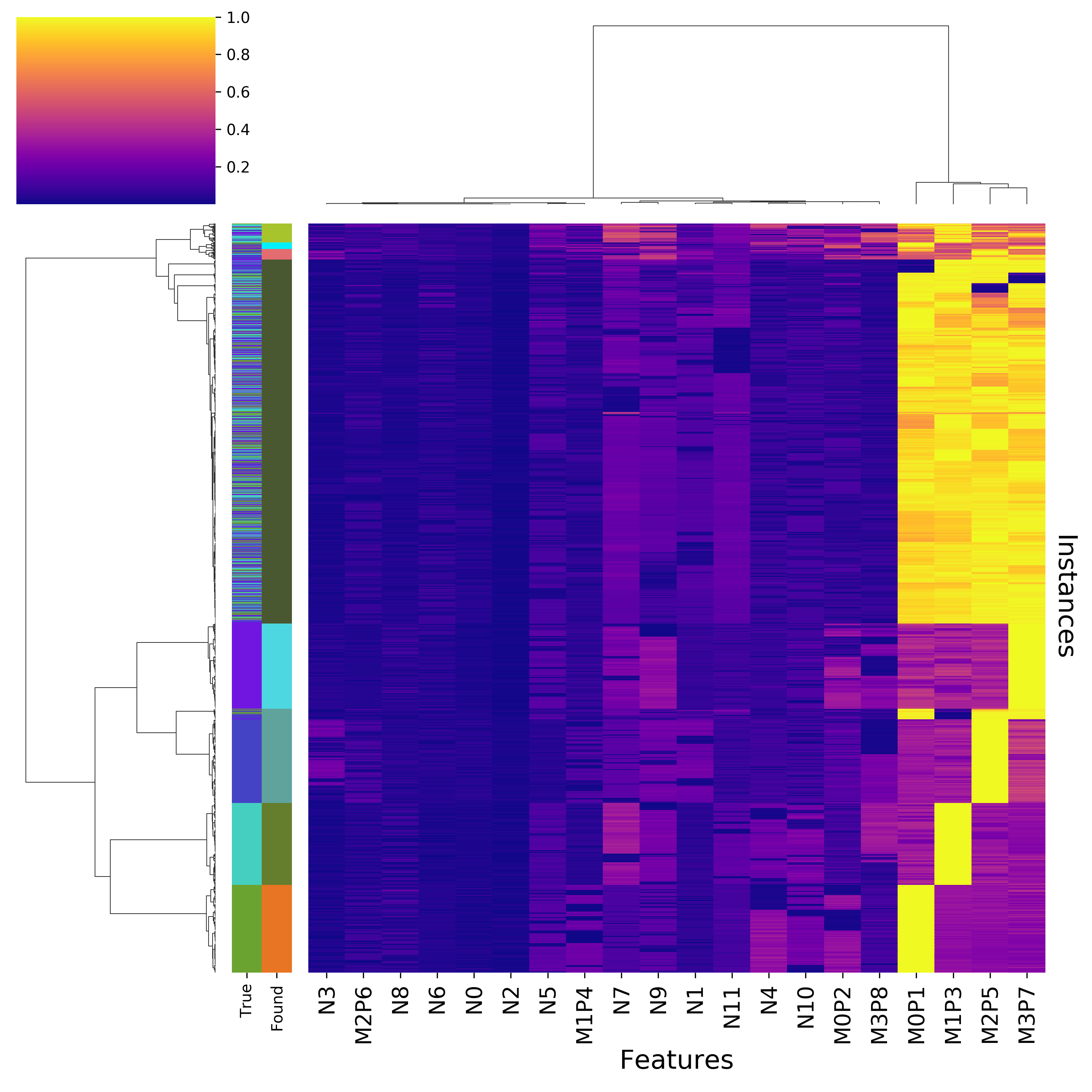}}}%
    \qquad
    \subfloat[\centering D18: Noisy 4-feature heterogeneous univariate]{{\includegraphics[width=0.4\linewidth, keepaspectratio]{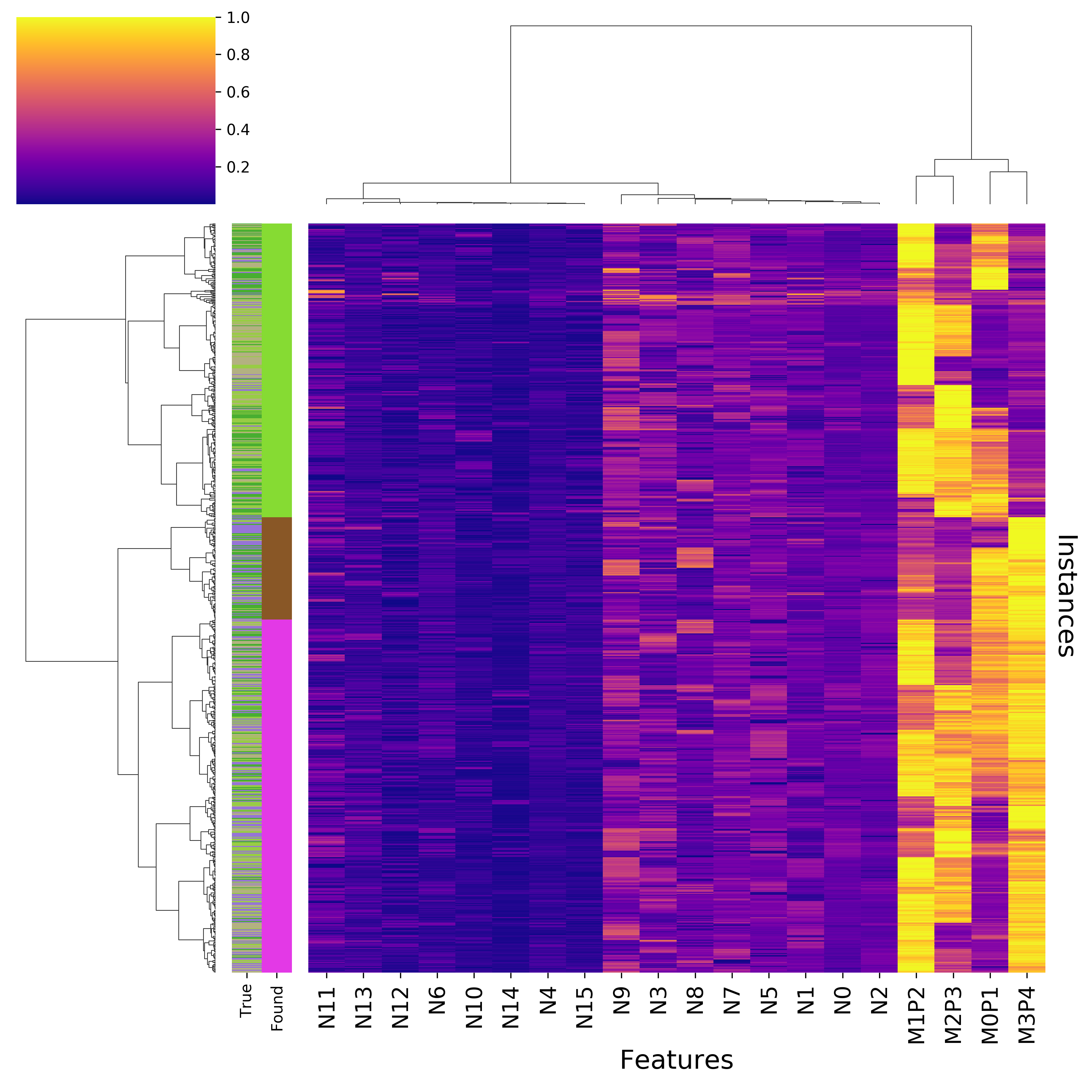}}}%
    \qquad
    \subfloat[\centering D19: Clean 2 sets of 2-way Epistasis, heterogeneous]{{\includegraphics[width=0.4\linewidth, keepaspectratio]{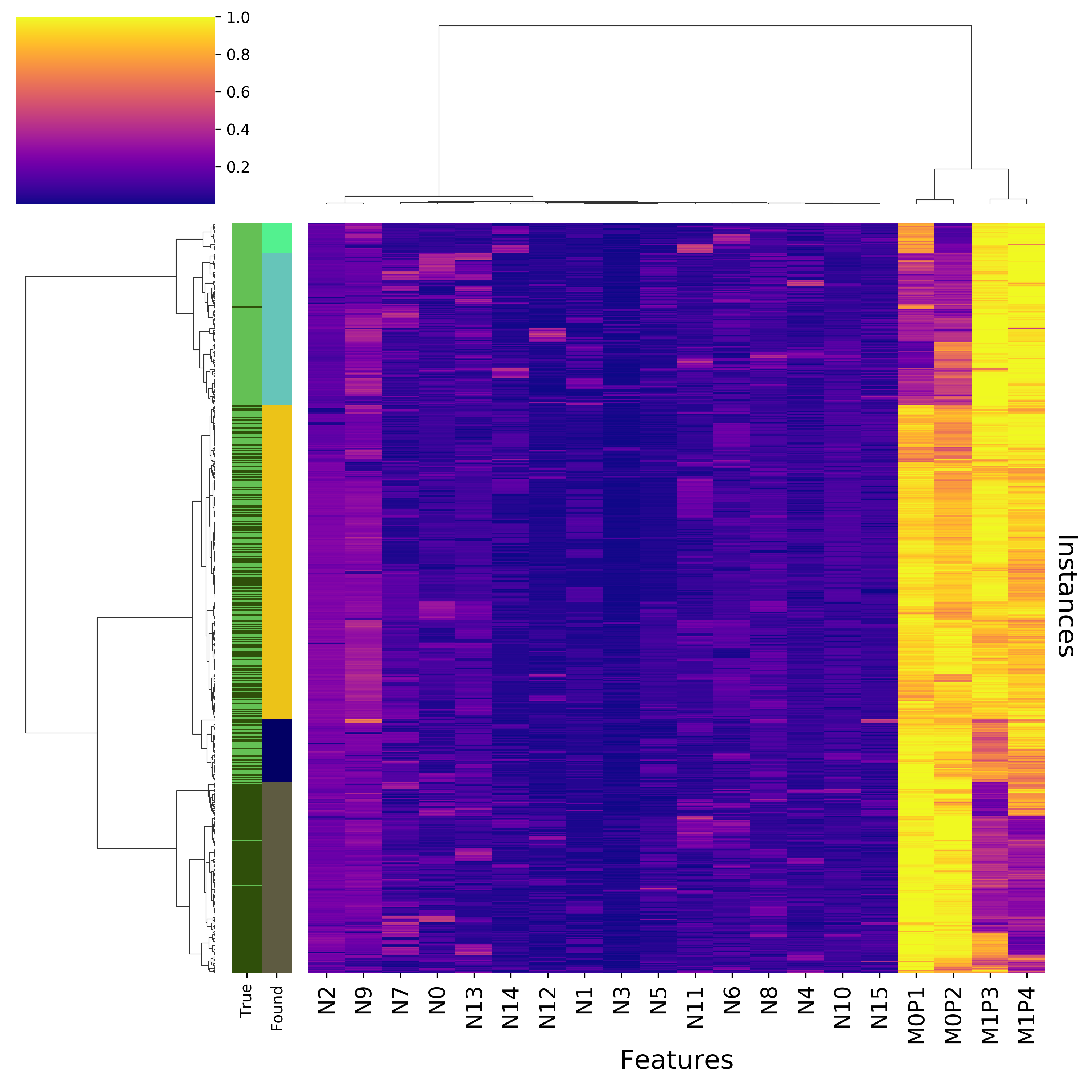}}}%
    \qquad
    \subfloat[\centering D20: Noisy 2 sets of 2-way Epistasis, heterogeneous]{{\includegraphics[width=0.4\linewidth, keepaspectratio]{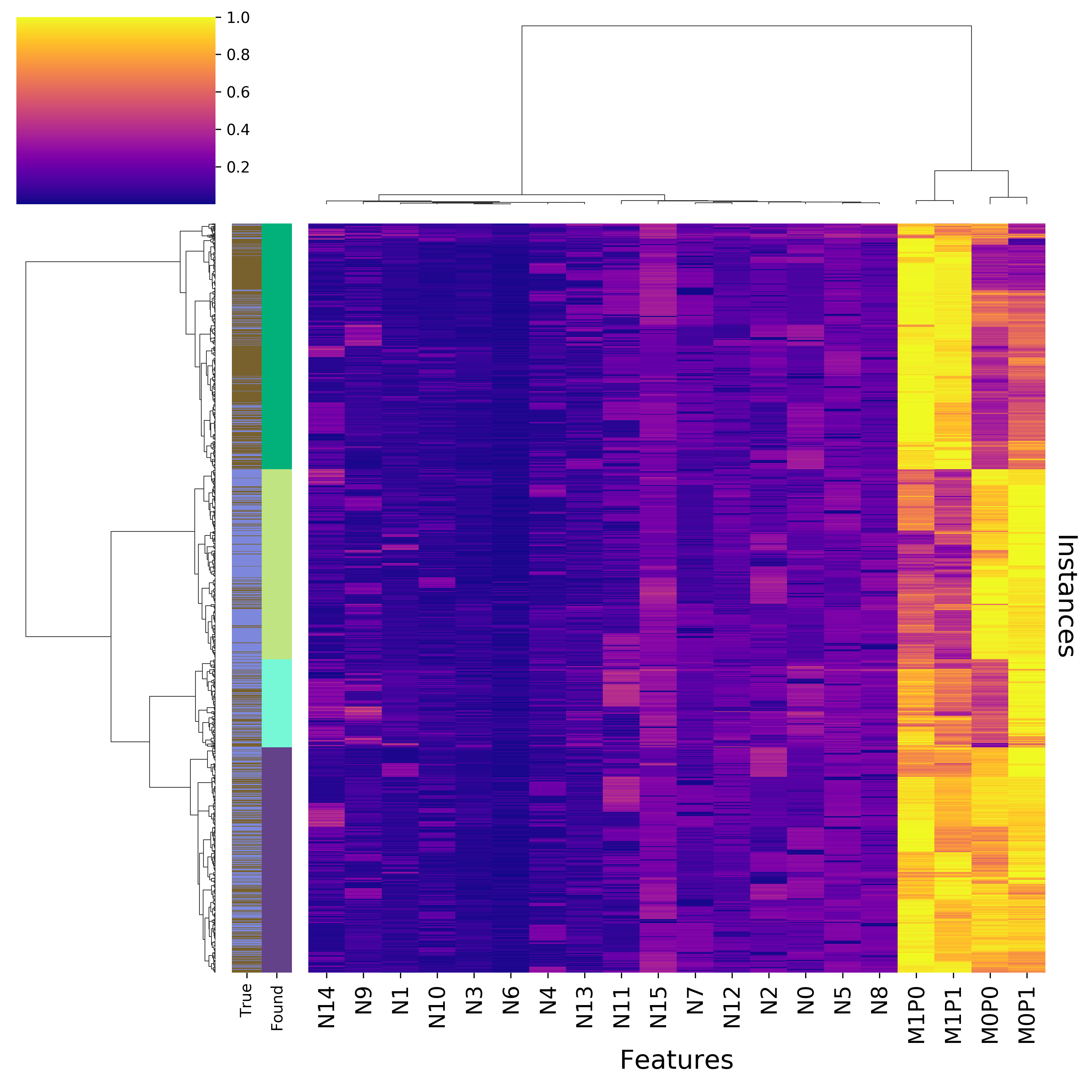}}}%
    \caption{FT Clustermaps of heterogeneous GAMETES datasets D15-D20}
    \label{fig:HetME}
\end{figure}

Here we examine clean and noisy GAMETES datasets D15-D21 examining scenarios with heterogeneous associations. FT clustermaps for heterogeneous combinations of associations are given in Figure \ref{fig:HetME}. Now that we are simulating underlying heterogeneous subgroups, these figures now include both the 'found' and 'true' clusters for reference. The number of found clusters presented in each plot is based on the automated recommendation.

For D15 (Figure \ref{fig:HetME}A), with 2 univariate associations combined heterogeneously, we observe three `found' clusters with strong unique signatures. Specifically, two of the clusters emphasize one of the two heterogeneous features and the third has both emphasized. Notice that the upper two found clusters completely overlap with underlying true clusters but the third found cluster is a mix of instances from either underlying model. This is an expected characteristic based on how heterogeneity is simulated with GAMETES. Specifically, instances are generated separately for each underlying model, then the instances are combined into a single dataset. Instances in the top cluster were generated with M0P1 being predictive, and the randomly generated feature value for M1P2 was not predictive in-line with the M1P2 model used to generate the other half of the data. Instances in the middle cluster followed the same pattern but with M0P1 and M1P2 switched. Lastly, in the lowest found cluster where both features were important, the instance feature values happened to correspond to correct classifications based on both the M0P1 and M1P2 models. This yields a signature for the third cluster that is more similar to an epistatic interaction. However this unique `overlapping' form of heterogeneity is characterized by the three distinct clusters. For D16 (Figure \ref{fig:HetME}B), we observe a very similar signature but with an additional cluster of noisy instances (second cluster from the top) which we confirm as noise by examining it's underlying cluster testing accuracy.

For D17 (Figure \ref{fig:HetME}C), with 4 univariate associations combined heterogeneously, we now observe 5 dominant `found' clusters, i.e. one where all 4 predictive features are important, and four others where only a single one is. The explanation for this expected finding is the same as for D15. For D18, (Figure \ref{fig:HetME}D) we observe the impact of 60$\%$ noise in combination with this `overlapping' form of heterogeneity on the clarity of the FT signature. The lack of a larger cluster where all predictive features have a high FT signature distinguishes it from D6 (i.e. noisy 4-feature additive). While LCS-DIVE has failed to clearly cluster the true underlying subgroups in this scenario, FT clearly distinguishes which features are relevant (i.e. predictive) from those that are not as in all other GAMETES analysis.

For D19 (Figure \ref{fig:HetME}E), we now consider both epistasis and heterogeneity simultaneously. Here we observe 3 dominant `found' clusters, i.e. where one 2-way epistatic interaction, the other, or both are important. This signature is consistent with what we observed for other clean GAMETES heterogeneity simulations, i.e. D15 and D17, but now also capturing underlying epistatic interactions relevant in distinct instance subsets. The extra `found' cluster are a reflection of the simulated model overlap as well as the model only reaching about 76 $\%$ testing accuracy within the allotted training iterations. For D20 (Figure \ref{fig:HetME}F), we observe the same, albeit messier, FT signature in comparison to D19, with three dominant `found' clusters. Again we confirm smaller clusters capture instances that were poorly predicted yielding a less consistent signature. Lastly, for D21 (see Figure 3 of Sup.3.4) we examine the FT signature when an unbalanced proportion of instances (i.e. 75:25) correspond with each of the heterogeneous epistatic models. In this scenario, the more frequent instance subgroup (i.e. 75$\%$) yields a clear 2-way interaction signature, while the second subgroup more ambiguously indicates the involvement of the two additional predictive features. This supports our finding that FT score signatures reliably identify key relevant features in the LCS model, and suggests that this approach will facilitate the characterization of dominant patterns of association while sometimes at the expense of clearly identifying a less frequent heterogeneous association.

\subsubsection{GAMETES Comparison Summary}
Looking across D1-D21 GAMETES simulation results, we found evidence that FT clustermap signatures are generally capable of distinguishing clean and noisy associations that are univariate, additive, epistatic, heterogeneous, and mixed. Further, interrogation of the testing accuracy within `found' clusters helps to discriminate heterogeneous subgroups from ignorable `noisy' clusters and thus guide the selection of the optimal number of clusters for downstream investigation. Another item to pay attention to is the testing accuracy of the LCS model as a whole. FT signature interpretation clearly relies on acknowledgment of how much signal was captured by the model, i.e. FT signatures uniquely vary in clean vs. noisy scenarios. Notably, the FT signatures of scenarios involving additive vs. heterogeneous signatures were similar enough to warrant further discussion and direct comparison. These include (D6 vs. D18) and (D14 vs.D20) as detailed in Sup.3.5. Also note that the overlapping simulation of heterogeneity (simulated by GAMETES) makes such associations more difficult to clearly characterize in contrast with MUX heterogeneity. Real world problems could exhibit heterogeneous associations of either type so we should be prepared to characterize them both. LCS-DIVE run time reports for all phases and datasets are given in Sup.3.6.

\subsection{LCS-DIVE Applied to Pancreatic Cancer Data}
We applied LCS-DIVE to the two pancreatic cancer datasets summarized in Table \ref{tab:realparams}. Phase 1 Scikit-ExSTraCS modeling yielded an average testing accuracy of $66.75\%$ and $68.2\%$ on P1 and P2 respectively. The addition of the dietary variables in P2 contributed to a slightly higher pancreatic cancer prediction accuracy.

Next, we turn to LCS-DIVE FT analysis (i.e. Phase 2). For P1, k = 53 significant candidate instance clusterings were identified from which 11 clusters was automatically recommended. Upon manual inspection we instead selected 4 optimal clusters for P1 based on subjective evaluation and the internal testing accuracies of candidate clusters. For P2, k = 18 significant clusters were found from which 4 clusters was recommended. Manual inspection confirmed these 4 clusters to be optimal. Elbow plots for P1 and P2 with automatic cluster recommendations are given in Sup.3.8.1. 

\begin{figure}[tph]
    \centering
    \subfloat[\centering P1]{{\includegraphics[width=0.8\linewidth, keepaspectratio]{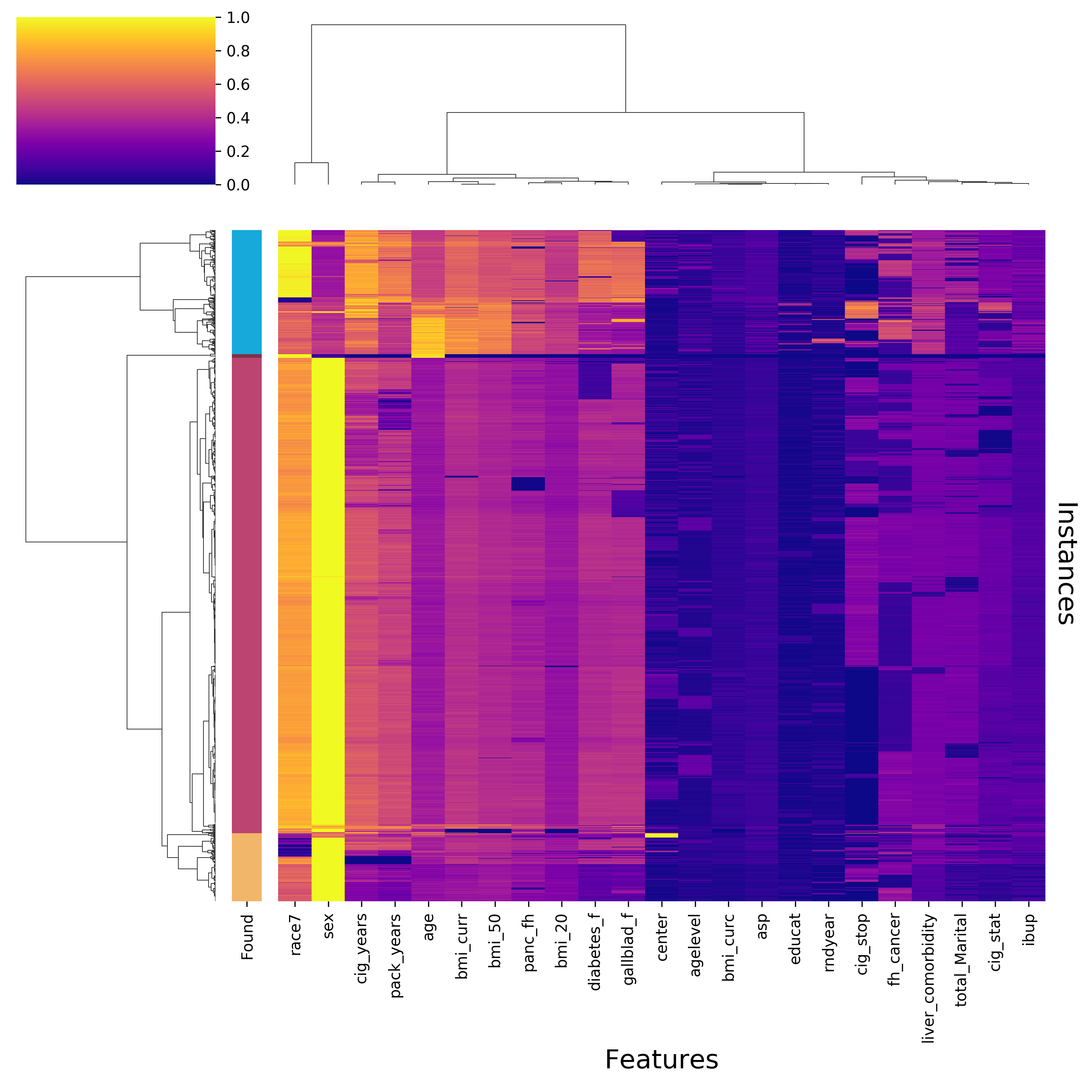}}}%
    \caption{P1 Pancreatic Cancer FT Clustermap}
    \label{fig:realplots1}
\end{figure}

Figure \ref{fig:realplots1} gives the FT clustermap for P1 with 4 `found' clusters chosen. With-cluster testing accuracies from the top cluster to bottom are as follows: 0.0287, 0.9667, 0.9789, and 0.9089. Thus, for FT signature evaluation we ignore the top cluster which captures instances that were not predicted accurately (i.e. noisy instances). In the second (very small) cluster, race and age alone drive accurate predictions. Further inspection of the instances in this cluster revealed missingness of many feature values outside of race and age explaining the very low FT signature for those features in this cluster. In the third (large) cluster, the covariates sex (in particular) and race, drove accurate predictions with a lesser (additive) importance signature for a number of other variables including: duration/years smoked, pack years of smoking, a family history of diabetes, and a family history of gallbladder or liver disease. Notably, this large cluster was > 99\% controls and 99.11\% of the male sex. In the bottom cluster, sex again dominates as being most predictive but all other variables were less predictive than in the large cluster. This cluster was 69\% cases and 63.18\% female. This may suggest a sex-driven heterogeneous effect where a number of other variables contribute to low-risk of pancreatic cancer in males (in the large cluster), but sex alone is the primary outcome predictor in the bottom cluster. This is not intended as a definitive finding but to illustrate how FT clustermaps can be leveraged to guide interpretation of the model as well as generate new hypotheses to guide downstream analyses. 

The P2 dataset FT clustermap (given in Sup.3.7.1) yielded similar high testing accuracy clusters to the bottom two clusters found in P1, however in this case, a number of the added dietary variables yielded the strongest secondary importance signatures after sex and race, in particular: total daily intake of carbohydrates, total fat, and total calcium. Given that the other lesser importance variables mentioned for P1 above no longer stood out in the P2 FT clustermap, it suggests that these dietary variables were more informative, also supported by the observed increase in model prediction accuracy. Rule-population visualizations (i.e. clustermap and network) for P2 were also generated to illustrate Phases 3 and 4 of LCS-DIVE in a real-world problem (see Sup.3.7.2). While useful to further interrogate the underlying rule-based model, we found FT clustermaps to be far more reliable and informative. Overall, this real-world application of LCS-DIVE illustrates the potential of this approach to guide interpretation of rule-based ML modeling and instance subgroup characterization.

\section{Conclusions}
This work has introduced and evaluated LCS-DIVE as an automated framework to characterize and distinguish underlying patterns of association detected by ExSTraCS, a rule-based ML algorithm designed to address complex, noisy, and heterogeneous biomedical classification. Further, we implemented scikit-ExSTraCS, a scikit-learn compatible implementation to facilitate application of rule-based ML and its comparison to other established ML approaches. The primary focus in this study was to determine the ability of LCS-DIVE to distinguish univariate, additive, epistatic, heterogeneous, and mixed underlying associations using FT score signatures in clean and noisy data. This was done through application to benchmark MUX datasets, GAMETES simulated SNP datasets, and a deeper dive into a real-world investigation of pancreatic cancer. 

Overall, LCS-DIVE yielded clearly interpretable signatures for clean MUX and GAMETES datasets, as well as largely reliable signatures for noisy simulated scenarios. This demonstrates the potential of FT signatures to provide insights for LCS model interpretation with respect to both feature importance and characterizing underlying patterns of association. We demonstrated how the automated clustering procedure can successfully recapitulate underlying heterogeneous subgroups. We also demonstrated how followup analysis of identified clusters can differentiate candidate heterogeneous subgroups from ignorable clusters in noisy problems. Compared to the more conventional rule-population analysis, FT score analysis produced a much cleaner, nuanced visualization. Application of LCS-DIVE to an investigation of pancreatic cancer suggested a possible sex-based heterogeneous association as well as a variety of multivariate additive effects. This work highlights the unique potential of rule-based ML approaches to not only detect complex associations but provide a strategy for global model interpretation and discovery of feature-driven instance subgroups in the presence of heterogeneity. 

Future work will seek to apply this automated pipeline to a broader selection of real-world datasets, and improve it's scalability. We will also seek to expand this pipeline to automatically conduct within-cluster analyses to further facilitate disambiguation of similar FT signatures in the special cases identified in this work (e.g. additive vs. heterogeneous feature combinations).

\section*{Acknowledgements}
This work was supported by a Department of Defense (DOD) Career Development Award-W81XWH-17-1-0276 (Dr. Shannon M. Lynch), the Intramural Research Program, Division of Cancer Epidemiology and Genetics, NCI (Dr. Rachael Stolzenberg-Solomonas), the National Institutes of Health Grant R01 HL134015 (Approaches to Genetic Heterogeneity of Obstructive Sleep Apnea), and a CURF award for Faculty Mentoring of Undergraduate Research at the University of Pennsylvania.

\bibliographystyle{abbrv}
\bibliography{ml_pipe.bib}  




\end{document}